\pdfoutput=1

\documentclass[11pt]{article}

\usepackage[final]{acl}

\usepackage{times}
\usepackage{latexsym}
\usepackage[T1]{fontenc}
\usepackage[utf8]{inputenc}
\usepackage{microtype}
\usepackage{inconsolata}
\usepackage{graphicx}

\newcommand{\dataset}{ToM-SSI}
\usepackage{subcaption} 
\usepackage{microtype}
\usepackage{booktabs}
\usepackage{amsmath}
\usepackage{amssymb}
\usepackage[dvipsnames]{xcolor}
\usepackage[most]{tcolorbox}
\usepackage{listings}
\tcbuselibrary{listings}
\usepackage{enumitem}
\usepackage[table]{xcolor}
\usepackage{algorithm}
\usepackage{algpseudocode}
\usepackage{multirow}

\newcommand*{\img}[1]{%
    \raisebox{-.2\baselineskip}{%
        \includegraphics[
        height=\baselineskip,
        width=\baselineskip,
        keepaspectratio,
        ]{#1}%
    }%
}

\tcbset{
    myboxstyle/.style={
        enhanced,
        breakable,
        width=\textwidth,
        listing only,
    }
}

\newtcolorbox[auto counter]{example}[3]{
    myboxstyle,
    label=#1,
    title=Example~\thetcbcounter:~#3,
    colback=#2!5!white,
    colframe=#2!75!black,
}

\newcommand{\red}{\img{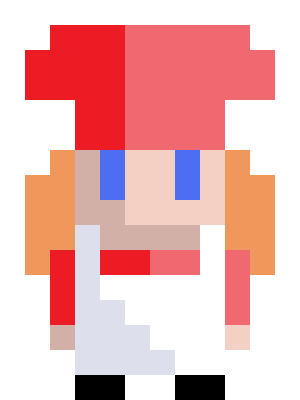}}
\newcommand{\blue}{\img{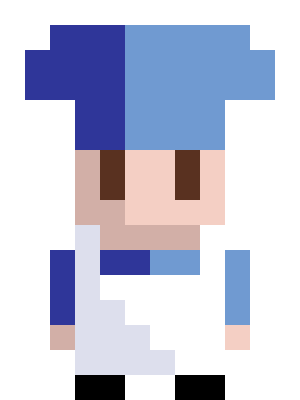}}
\newcommand{\green}{\img{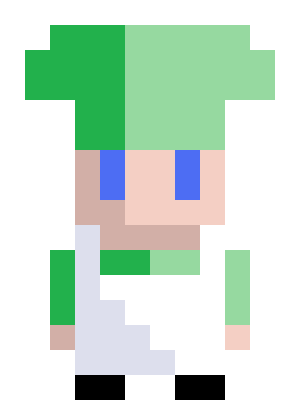}}
\newcommand{\violet}{\img{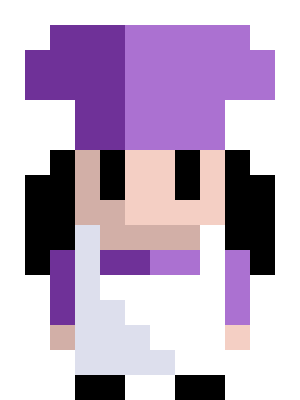}}
\newcommand{\bread}{\img{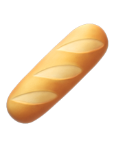}}
\newcommand{\salad}{\img{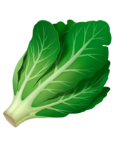}}
\newcommand{\tomato}{\img{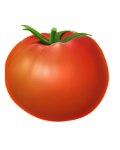}}
\newcommand{\meat}{\img{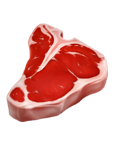}}
\newcommand{\blank}{\img{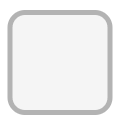}}

\definecolor{a0}{RGB}{234,51,35}
\definecolor{a1}{RGB}{0,0,245}
\definecolor{a2}{RGB}{55,126,33}
\definecolor{a3}{RGB}{117,20,124}
\newcommand{\azero}{\textcolor{a0}{$A_0$}}
\newcommand{\aone}{\textcolor{a1}{$A_1$}}
\newcommand{\atwo}{\textcolor{a2}{$A_2$}}
\newcommand{\athree}{\textcolor{a3}{$A_3$}}

\definecolor{PerceptColor}{HTML}{BFB299}
\definecolor{BeliefColor}{HTML}{D4A27F}
\definecolor{IntentColor}{HTML}{CC785C}
\definecolor{DesireColor}{HTML}{999999}
\definecolor{HearColor}{RGB}{228,235,245}
\newcommand\Pcolor[1]{\textcolor{PerceptColor}{#1}}
\newcommand\Bcolor[1]{\textcolor{BeliefColor}{#1}}
\newcommand\Icolor[1]{\textcolor{IntentColor}{#1}}

\definecolor{PerceptBoxColor}{HTML}{EBDBBC}
\definecolor{BeliefBoxColor}{HTML}{FFC399}
\definecolor{IntentBoxColor}{HTML}{FFAB85}
\newcommand\Pboxcolor[1]{\tcbox[on line, boxsep=1.5pt, left=0pt,right=0pt,top=1pt,bottom=0pt,colframe=white,colback=PerceptBoxColor,highlight math style={enhanced}]{#1}}
\newcommand\Bboxcolor[1]{\tcbox[on line, boxsep=1.5pt, left=0pt,right=0pt,top=0pt,bottom=2pt,colframe=white,colback=BeliefBoxColor,highlight math style={enhanced}]{#1}}
\newcommand\Iboxcolor[1]{\tcbox[on line, boxsep=1.5pt, left=0pt,right=0pt,top=0pt,bottom=2pt,colframe=white,colback=IntentBoxColor,highlight math style={enhanced}]{#1}}

\title{ToM-SSI: Evaluating Theory of Mind in Situated Social Interactions}

\author{
Matteo Bortoletto \quad 
Constantin Ruhdorfer \quad 
Andreas Bulling \\
University of Stuttgart, Germany \\
\texttt{matteo.bortoletto@vis.uni-stuttgart.de} 
}

\begin{document}

\everypar{\looseness=-1}

\maketitle

\begin{abstract}
    Most existing Theory of Mind (ToM) benchmarks for foundation models rely on variations of the Sally-Anne test, offering only a very limited perspective on ToM and neglecting the complexity of human social interactions.
To address this gap, we propose \dataset{}: a new benchmark specifically designed to test ToM capabilities in environments rich with \textit{social interactions and spatial dynamics}. 
While current ToM benchmarks are limited to text-only or dyadic interactions, \dataset{} is multimodal and includes group interactions of up to four agents that communicate and move in \textit{situated} environments.
This unique design allows us to study, for the first time, mixed cooperative-obstructive settings and reasoning about multiple agents' mental state in parallel, thus capturing a wider range of social cognition than existing benchmarks.
Our evaluations reveal that the current models' performance is still severely limited, especially in these new tasks, highlighting critical gaps for future research.
\footnote{\img{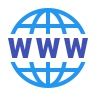} Project web page: \url{https://collaborative-ai.org/publications/bortoletto25_emnlp/}.}
\end{abstract}

\section{Introduction}

Theory of Mind (ToM) is the ability to attribute mental states to oneself and others, such as beliefs, intents, desires, or knowledge \citep{premack1978does}. 
ToM is important in human social interactions as well as for empathy and effective communication, all of which are inherently grounded in a physical environment.

Recent advances in large foundation models (LFMs) have spurred the creation of benchmarks to assess LFMs' ToM abilities, but these benchmarks suffer from important limitations.
\begin{figure}[t]
    \centering
    \includegraphics[width=\linewidth]{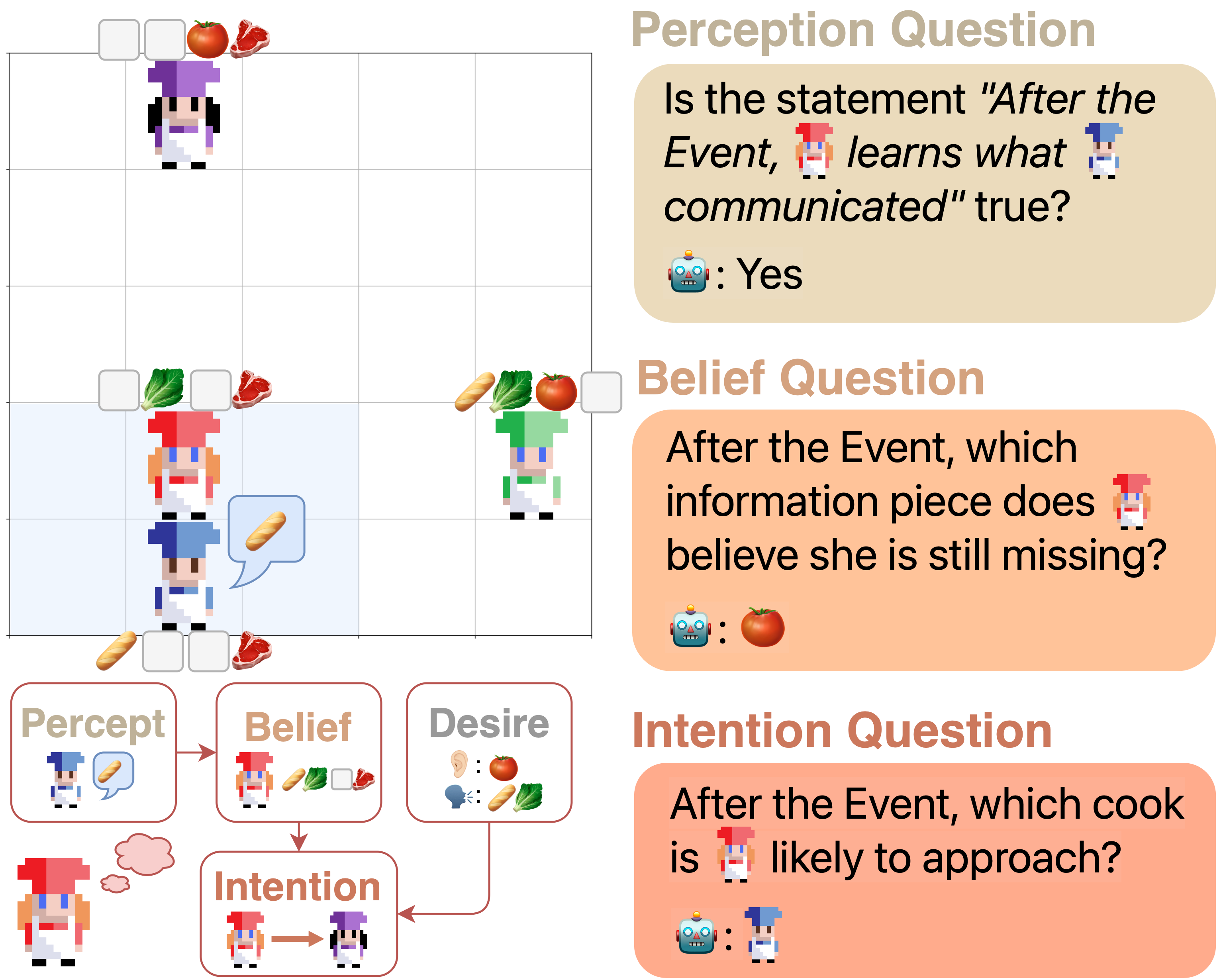}
    \caption{\dataset{} is both physically and socially situated, introducing up to four agents moving and communicating in a grid world environment with the goal of sharing and acquiring information. Since agents possess asymmetric information and communication is \colorbox{HearColor}{spatially constrained}, \dataset{} requires models to take their perspective to reason about their perceptions, beliefs, desires, and intentions.}
    \label{fig:teaser}
\end{figure}
Many benchmarks ~\citep{le2019revisiting, sclar2023minding, ma2023tomchallenges, he2023hi, gandhi2024understanding, xu2024opentom, zhou2023far} are based on variations of the prototypical Sally-Anne test \citep{wimmer1983beliefs}, where Sally places an object in a location, leaves, and Anne moves it.
An observer is then tested to see whether they can understand that Sally will hold a false belief about the object's location upon her return.
Despite its popularity, the Sally-Anne test only offers a limited perspective on ToM, and it neglects the complexity of social interactions.
While other benchmarks cover a broader range of social interactions, they still only involve textual input \citep{kim2023fantom, chen2024tombench, hou2024entering, gu2024simpletom}.

ToM evaluations must be both physically situated -- requiring the interpretation of visual cues or spatial relationships -- and socially situated in interactions between multiple agents \cite{ma2023towards}.
Most recent benchmarks try to address both limitations by using simulated environments \cite{bara2021mindcraft, bara2023towards, bortoletto2024limits, jin2024mmtom, shi2024muma}, but they still only consider interactions between two agents.
As a result, they are limited to simple cooperative or obstructive tasks and require models to track the mental states of at most two agents.

We introduce \dataset{}, the first evaluation benchmark that addresses all aforementioned limitations by evaluating \textbf{ToM} abilities in \textbf{S}ituated \textbf{S}ocial \textbf{I}nteractions:
\begin{enumerate}[leftmargin=15pt,topsep=5pt,itemsep=1pt,partopsep=1ex,parsep=1ex]
    \item \dataset{} goes beyond the Sally-Anne test by covering agents that move and communicate in a rich social environment to share and acquire different pieces of information.
    \item \dataset{} is formulated as a visual-text question answering task \citep{chen2024vtqa} and is thus inherently multimodal.
    Since agents possess asymmetric knowledge and their communication is spatially constrained, \dataset{} requires LFMs to align spatial information in images with textual descriptions and to take agents' perspective to reason about their percepts, beliefs, desires, and intentions (see Figure~\ref{fig:teaser}).
    \item \dataset{} supports triadic and tetradic social interactions, allowing us to evaluate different agent attitudes.
    It comprises five tasks involving cooperative movement, cooperative and obstructive communication, and mixed cooperative-obstructive communication (see Figure~\ref{fig:tomssi}) -- featuring 6,000 questions in total. 
\end{enumerate}

We report evaluations using \dataset{} that reveal several important and novel insights. 
We demonstrate that current state-of-the-art LFMs perform significantly worse than humans.
For certain tasks, they even perform worse than smaller models. 
We further show that models struggle with two critical steps necessary for reasoning about agents’ percepts, beliefs, and intentions: (1) inferring the percepts of a target agent, and (2) determining that agent's beliefs based on those percepts. 
We then analyse error cases in two challenging \dataset{} tasks that reveal the limitations of models in tracking nested beliefs in multi-agent communication and modelling mixed social interactions. 
Overall, our evaluations show that current models’ ToM abilities are still severely limited, particularly in the new tasks introduced by \dataset{}, highlighting critical gaps for future research.

\section{\dataset{}}

\subsection{Designing Situated Social Interactions}
\label{sec:design}

\paragraph{Grid World}
As in prior work \citep{rabinowitz2018machine, Sclar2022Symmetric, gandhi2021baby}, we opted for a grid world environment where all agent interactions occur.
A grid world allows us to study the core abilities targeted by \dataset{}, while minimising complexities that could compromise assessment clarity -- such as hallucinations~\citep{sahoo2024comprehensive}.
We create different grid layouts by applying geometric transformations (see \ref{app:transform}) to minimal templates inspired by previous work in deep reinforcement learning~\citep{Sclar2022Symmetric}, where up to four agents are placed in pre-defined locations.
The grid world is then rendered as an image. 
To be able to evaluate models that only support text as input, we also generate a character version of each grid, as shown in Figure~\ref{fig:grid}. 
Both grid versions retain the same information required for performing the task. 

\paragraph{Agents} 
Each agent $A_j$, $j\in[0, 3]$, occupies one cell in the grid and starts with partial knowledge $I_{A_j} \subseteq I = \{i_0, i_1, i_2, i_3\}$. 
Agents have two goals: 1) fill their knowledge gaps by learning missing pieces of information from other agents, and 2) share their knowledge with other agents who lack that information.
While agents know the initial positions, movements, and starting knowledge of others, they cannot directly see new information that other agents may acquire later.
Instead, they must infer it from events that they observe. 
Communication is spatially constrained: an agent $A_j$ can lean information that another agent $A_k$ is communicating only if $A_j$ is in one of the adjacent cells to $A_k$, i.e.\ 
$(x_j, y_j) \in \{(x, y) : 1 \leq |x - x_k| + |y - y_k| \leq 2 \}$,
where $(x_j, y_j)$ and $(x_k, y_k)$ represent the grid coordinates of agents $A_j$ and $A_k$, respectively.
For example, in Figure~\ref{fig:teaser}, only \red{} can learn what \blue{} is communicating.

\begin{figure*}[t]
    \centering
    \includegraphics[width=\textwidth]{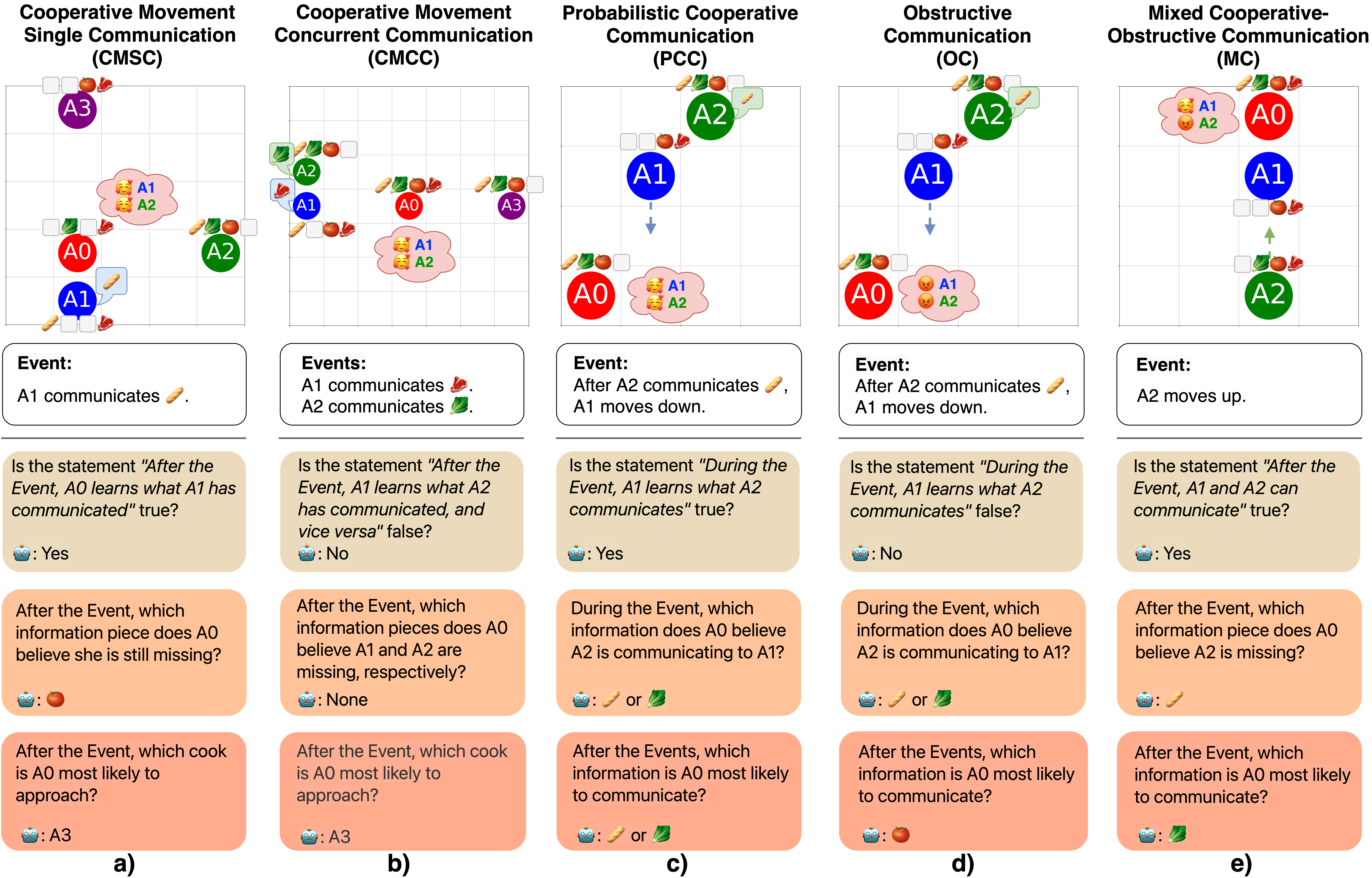}
    \caption{\textbf{Overview of \dataset{}.} \dataset{} extends to triadic and tetradic social interactions, covering different agent attitudes: cooperative, obstructive, and mixed settings. The dataset comprises five tasks involving cooperative movement, cooperative communication, obstructive communication, and mixed cooperative-obstructive communication. Each sample is paired with three questions, covering agents' \Pboxcolor{percepts}, \Bboxcolor{beliefs}, and \Iboxcolor{intentions}.}
    \label{fig:tomssi}
\end{figure*}

A key novelty of \dataset{} is that it allows us to study a wider range of agent desires than previous benchmarks.
By default, agents have a \textit{collaborative} attitude, i.e., they want to share their knowledge with other agents.
\dataset{} also offers to design agents with an \textit{obstructive} attitude, who aim to prevent other agents from learning new information (see Figure~\ref{fig:tomssi}d).
Moreover, given that \dataset{} supports more than two agents, we can study \textit{mixed collaborative-obstructive} scenarios in triadic interactions in which one agent $A_j$ is collaborative with $A_k$ but obstructive towards $A_l$ (see Figure~\ref{fig:tomssi}e).

\paragraph{Events} 
Events dictate the flow of information and change the state of the grid, creating dynamic opportunities for collaboration and inference.
Events can involve agents' movement, communication of information, or both. 
Agents can move up, down, left, right, and/or communicate one piece of information they possess. 
For example, the event in Figure~\ref{fig:teaser} is ``Cook \blue{} communicates \bread''.

\paragraph{Social Context} 
Tasks in \dataset{} are situated in social contexts.
For example, the social context in Figure~\ref{fig:teaser} could be ``a restaurant's kitchen in which four chefs are preparing a dish'' or ``a cooking class where four participants are learning to make a new dish''.
The four pieces of information are assigned to different IDs according to the context, for example, the ingredients needed to make the dish: ``bread'' (\bread), ``salad'' (\salad), ``tomato'' (\tomato), and ``meat'' (\meat).
We generated a collection of 121 social contexts using GPT-4o \citep{gpt4o} and randomly sampled from this collection while generating \dataset{} samples (details in \ref{app:stories}).

\subsection{Question Types}
Inspired by the Belief-Desire-Intention framework \cite{bratman1987intention, baker2011bayesian}, samples in \dataset{} are paired with three questions, covering agents' \textit{percept}, \textit{beliefs}, and \textit{intentions}, given their \textit{desires}, as shown in Figure \ref{fig:teaser}. 
We do not include questions about \textit{desires} as they trigger all social interactions and are already specified in the textual prompt.
Given the percept-belief-intention causal structure shown in Figure~\ref{fig:teaser}, percept questions act as a control for belief questions (\Pcolor{Percept} $\to$ \Bcolor{Belief}), which in turn serve as a control for intention questions (\Bcolor{Belief} $\to$ \Icolor{Intention}). 
Therefore, a model with strong ToM abilities must answer all three questions correctly.

\paragraph{\Pcolor{Percept}}
Percepts are the observations an agent makes about the environment, forming its understanding of the current world state. 
In \dataset{}, percepts include the agents' positions, starting knowledge, and movement.
Percept questions probe whether models can accurately attribute agents to percepts based on the information presented in the image and text.
This ability is called \textit{perspective taking} -- a foundational ability in ToM \cite{masangkay1974early}.
For example, in Figure~\ref{fig:teaser}, the percept question asks whether \red{} learns what \blue{} has communicated.
To answer correctly, one must observe that the two agents occupy adjacent cells so \red{} can hear what \blue{} communicates.
Percept questions are framed as yes/no questions in the form ``Is the statement [...] true/false?''.

\paragraph{\Bcolor{Belief}}
Beliefs are an agent's internal representation of the world, derived from its percepts and prior knowledge \citep{perner1994understanding}. 
Beliefs may include assumptions about hidden aspects of the world, such as what another agent knows.
Belief questions evaluate whether models understand what information an agent knows or is communicating based on the agent's percepts.
For example, in Figure~\ref{fig:teaser}, models must infer that because \red{} can hear what \blue{} communicates, \red{} will believe she is missing only one piece of information (\tomato).
Belief questions are multiple-choice questions with the information IDs as options.

\paragraph{\Icolor{Intention}}
Intentions are the specific plans or actions that, given their beliefs and desires, an agent commits to achieve their goals \citep{tomasello2005understanding}. 
Intention questions focus on whether models can deduce the agents' action -- either communicative or motor.
For example, in Figure~\ref{fig:teaser}, models must infer that, given her desires (learning and sharing information) and her belief (missing \tomato), \red{} is more likely to approach \violet\nolinebreak. 
This is because \violet{} can communicate the last piece of information that \red{} is missing (\tomato), and \red{} can communicate the two pieces of information that \violet{} is missing (\bread and \salad).
In comparison, approaching \green{} would be suboptimal, as \red{} could only share one piece of information that \green{} lacks (\meat). 
Intention questions are multiple-choice questions, with options being agent IDs in case of movement and information IDs in case of communication. 

\subsection{Tasks}
\label{sec:tasks}
\dataset{} comprises 6,000 questions, equally split between five tasks.
These tasks reflect different aspects of everyday social interactions involving agents with collaborative, obstructive, or mixed attitudes. 
They require tracking multiple agents' beliefs, interpreting various communication events, and making inferences under uncertainty.

ToM tasks must satisfy two important criteria~\citep{quesque2020theory}.
First, they must require models to differentiate between mental states (in our case, knowledge) of different agents (\textit{non-merging criterion}).  
Second, it should be impossible to pass the tasks using low-level heuristics (\textit{mentalising criterion}).
Tasks in \dataset{} fulfil these criteria by requiring models to integrate spatial information from the grid with events and agents' knowledge. 
Given that agents have partial observability of communicative events, tasks in \dataset{} also require models to take agents' perspectives when answering questions.

We discuss how to solve each task based on formal utility functions in \ref{app:utilityf}.
For simplicity, in the following task descriptions, all questions only target \azero{} and we use the cooking example of Figure~\ref{fig:teaser} to label the information IDs. 
Figure~\ref{fig:tomssi} illustrates examples from the five tasks, where we show original images from the dataset with overlapping knowledge and attitude.

\subsubsection{Cooperative Movement -- Single Communication (CMSC)}
\label{subsub:coop-act-single}
The CMSC task requires models to reason about one agent's mental state as that agent is involved in a communicative action.
The task includes four agents, \azero{}, \aone{}, \atwo{}, and \athree{}, as shown in Figure~\ref{fig:tomssi}a. 
During the event, agent \aone{} communicates a piece of information (\bread) that agent \azero{}{} is missing and able to learn.

\paragraph{Desire} \azero{} wants to learn new information from other agents and share information that other agents are missing. 

\paragraph{Percept}  
The percept question asks if, after the event, \azero{} learns what \aone{} communicated. 
This serves as a control to verify whether a model can observe that \azero{} and \aone{} occupy adjacent cells on the grid. 
Recognising adjacency is crucial for inferring \azero{}'s belief and intention.

\paragraph{Belief}  
The belief question asks to identify which information \azero{} believes she is still lacking. 
Initially, \azero{} knows \salad{} and \meat.
Upon learning \bread{} from \aone{}, the model must infer that the only remaining missing information for \azero{} is \tomato.

\paragraph{Intention}  
The intention question asks who is \azero{} most likely to approach next. 
After the event, \azero{} knows \bread, \salad, and \meat. 
A model must correctly infer that \azero{} will move toward \athree{}, as \athree{} can provide the last missing piece (\tomato), and \azero{} can share the two pieces of information that \athree{} is missing.  
A model that incorrectly infers that \azero{} does not learn \bread{} will likely predict that \azero{} would go to \atwo{} instead (since \atwo{} knows both \bread{} and \tomato, and \azero{} can still communicate \meat).

\subsubsection{Cooperative Movement -- Concurrent Communication (CMCC)}
\label{subsub:coop-act-conc}

The CMCC task differs from the CMSC task in that
it challenges models to reason about one agent’s mental states while this agent observes two other agents communicating with each other at the same time (Figure~\ref{fig:tomssi}b). 
However, the observing agent cannot know for sure what is being communicated because she is too far away to hear them.
During the event, agents \aone{} and \atwo{}, who occupy adjacent cells, share the information that each is missing.

\paragraph{Desire} \azero{} wants to learn new information from other agents and share information that other agents are missing. 

\paragraph{Percept} 
The percept question asks if, after the event, \aone{} learns what \atwo{} has communicated, and vice versa. 
This serves as a control to verify whether a model can observe that \aone{} and \atwo{} occupy adjacent cells on the grid, and therefore can learn from each other. 

\paragraph{Belief} 
The belief question asks to identify which information \azero{} believes \aone{} and \atwo{} are still lacking.  
A model needs to infer that \azero{} will believe that since \aone{} and \atwo{} possess what the other is missing and can communicate, after the event they will possess all the information.

\paragraph{Intention}
The intention question asks who is \azero{} most likely to approach next.  
A model that correctly infers \azero{}'s belief that after the event \aone{} and \atwo{} possess all the information will predict that the only sensible agent that \azero{} can approach is \athree{}, to share the only piece of information \athree{} is missing (\meat).

\subsubsection{Probabilistic Cooperative Communication (PCC)} 
\label{subsub:prob-coop-comm}

Tasks in PCC are set in a \textit{probabilistic} scenario, as the target agent's intention is subject to uncertainty. 
The task includes three agents \azero{}, \aone{}, and \atwo{}, as shown in Figure~\ref{fig:tomssi}c. 
In the event, agent \atwo{} communicates one of the two pieces of information \aone{} is missing (\bread{} or \salad, randomly chosen). 
Then, \aone{} moves adjacent to \azero{}. 

\paragraph{Desire} \azero{} wants to learn new information from other agents and share information that other agents are missing. 

\paragraph{Percept}
The percept question asks if, during the event, \aone{} learns what \atwo{} has communicated. 
This verifies whether a model can observe that \aone{} and \atwo{} occupy adjacent cells, and therefore \aone{} can learn what \atwo{} is communicating.

\paragraph{Belief} 
The belief question asks which information \azero{} believes \atwo{} has communicated to \aone{} during the event.   
Even if the event specifies which information \atwo{} is communicating, a model must infer that \azero{} will not know which specific information was communicated by \atwo{}. 
However, \azero{} will rationally believe it to be one of the pieces that \aone{} is missing (either \bread{} or \salad).

\paragraph{Intention}
The intention question asks which information \azero{} will most likely communicate after the event.  
A model that correctly infers \azero{}'s belief that \atwo{} is likely communicating one of the two pieces of information \aone{} is missing will predict that \azero{} will likely communicate one of these two, aware of the uncertainty of potentially communicating the same information that \aone{} has already learnt from \atwo{}. 

\subsubsection{Obstructive Communication (OC)} 
\label{subsub:obstr-comm}

The OC task is perceptually identical to PCC (\S\ref{subsub:prob-coop-comm}), but in this case, the target agent has an \textit{obstructive} attitude (Figure~\ref{fig:tomssi}d). 
Percept and belief questions are analogous to \S\ref{subsub:prob-coop-comm}. 

\paragraph{Desire} \azero{} does not want other agents to learn new information. 

\paragraph{Intention} 
The intention question asks which information \azero{} will most likely communicate after the event.  
Given \azero{}'s obstructive attitude, a model must predict that \azero{} will likely 
communicate the only piece of information that all the agents already know (\tomato).

\subsubsection{Mixed Cooperative-Obstructive Communication (MC)}
\label{subsub:mixed-comm}

The MC task extends to \textit{mixed cooperative-obstructive} settings, as the target agent is collaborative towards one agent but obstructive towards another. 
This task includes three agents \azero{}, \aone{}, and \atwo{}, as shown in Figure~\ref{fig:tomssi}e. 
In the event, agent \atwo{} moves one cell up. 

\paragraph{Desire} \azero{} wants \aone{} to gain new information while preventing \atwo{} from doing so.

\paragraph{Percept}
The percept question asks if, after the event, \aone{} and \atwo{} can communicate. 
This serves as a control to verify whether a model can combine information from the grid (\aone{} and \atwo{} are one cell apart) with information from the event (\atwo{} moving up).  
After the event, the two agents will occupy adjacent cells and, therefore, will be able to communicate.

\paragraph{Belief} 
The belief question asks which information \azero{} believes \atwo{} is missing.   
A model must infer that, based on \atwo{}'s initial knowledge and in the absence of further communications with other agents, \azero{} will believe that \atwo{} is missing \bread. 

\paragraph{Intention} 
The intention question asks which information \azero{} will most likely communicate after the event.  
Here, a model needs to infer that \azero{} believes that if she communicates \bread, \aone{} will learn it and then potentially pass it on to \atwo{}.  
However, this conflicts with \azero{}'s obstructive attitude towards \atwo{}. 
Instead, \azero{} is more likely to communicate \salad, which \aone{} is missing and \atwo{} already possesses.

\subsection{Utility Functions}
\label{app:utilityf}

Each of our tasks has one (or in some cases two, see Figure \ref{fig:tomssi}) correct answer(s) that can be formally found as discussed in \S\ref{sec:design}. 
It is also possible to infer the correct answer by defining the target agent's utility function for each task and maximising it.
Following the notation introduced in \S\ref{sec:design}, we can define the utility function for movement actions of agent $A_j$ as:
\begin{align*}
    U_{A_j}(\hat{I}_{A_k}) 
    &= U^{learn}_{A_j}(\hat{I}_{A_k}) + U^{share}_{A_j}(\hat{I}_{A_k}) \\
    &= |\hat{I}_{A_k} - I_{A_j}| + |I_{A_j} - \hat{I}_{A_k}|
\end{align*}
Where $\hat{I}_{A_k}$ denotes the information that $A_j$ believes is known by $A_k$, $|\hat{I}_{A_k} - I_{A_j}|$ is the number of pieces of information that agent $A_j$ can learn from agent $A_k$, and $|I_{A_j} - \hat{I}_{A_k}|$ is the number of pieces of information that agent $A_j$ can share with agent $A_k$. For example, in Figure 1: $U_{\tiny\red{}}(\small\violet{}) = 3$ and $U_{\tiny\red{}}(\small\green{}) = 2$. 
For communicative actions, we have three different cases:
\begin{itemize}
    \item Cooperative: $U_{A_j}(\hat{I}_{A_k}) = I_{A_j} - \hat{I}_{A_k}$
    \item Obstructive: $U_{A_j}(\hat{I}_{A_k}) = I_{A_j} \cap \hat{I}_{A_k}$
    \item Mixed, e.g. cooperative towards $A_k$ and obstructive towards $A_l$: \\$U_{A_j}(\hat{I}_{A_k}, \hat{I}_{A_l}) = (I_{A_j} - \hat{I}_{A_k}) \cap (I_{A_j} \cap \hat{I}_{A_l})$ 
\end{itemize}
Note that, in ToM-SSI, the spatial relationships are important to understand who is able to learn what, but we designed the tasks such that the number of steps required to reach another agent does not matter. 
Future versions of the benchmark could include rational movement as an additional complexity.

\subsection{Dataset Generation}

\dataset{} is entirely generated by code (see Algorithm \ref{alg:generation}). 
Each minimal template is paired with agents' initial knowledge and the correct answer, which is determined as discussed in \S\ref{sec:tasks} and \ref{app:utilityf}.
Starting from a minimal grid template, our generation pipeline applies random geometric transformations to the grid (see \ref{app:transform}) and samples one social context from our database to populate the prompt template corresponding to the task. 
To further avoid bias, agent and information IDs are randomly permuted.
We show the structure of prompts in \dataset{} in Figure~\ref{fig:prompt-structure}.
The prompt first introduces the social context and information about agents.
It then presents the grid with the agents in their initial position as an image (for VLMs) or text (for LMs), as shown in Figure~\ref{fig:grid}.
Following the grid, the prompt lists the information initially known by the agents, the attitude of the target agent, and the event(s) that trigger a change in the environment. 
The prompt ends with the question and multiple-choice answers for the model to select from.
To avoid bias, questions can have different formats (see \ref{app:prompts}).
Complete examples of prompts are included in \ref{app:prompts}.

\begin{figure*}[t]
    \centering
    \begin{subfigure}[b]{0.49\textwidth}
        \centering
        \includegraphics[width=\linewidth]{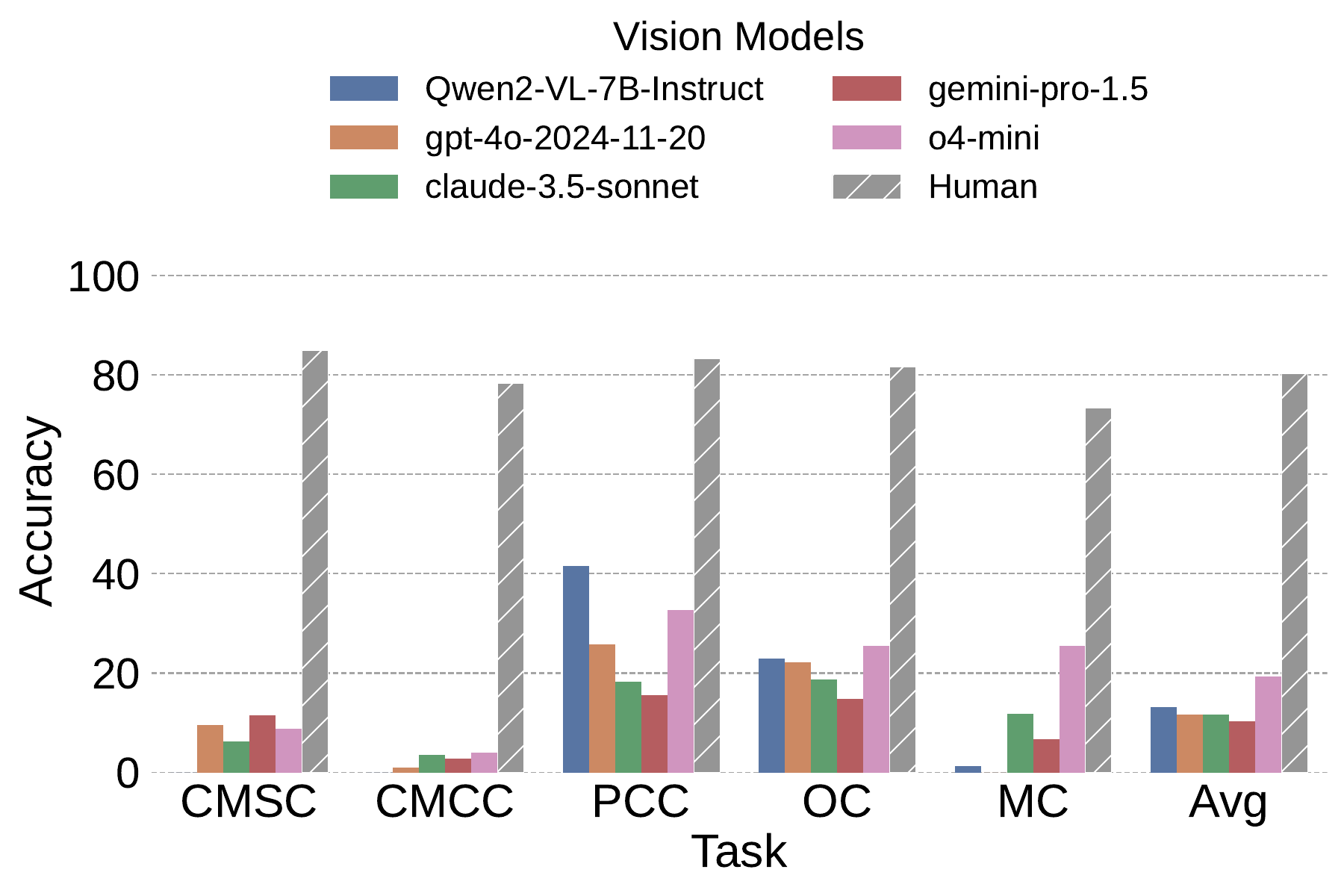}
    \end{subfigure}
    \begin{subfigure}[b]{0.49\textwidth}
        \centering
        \includegraphics[width=\linewidth]{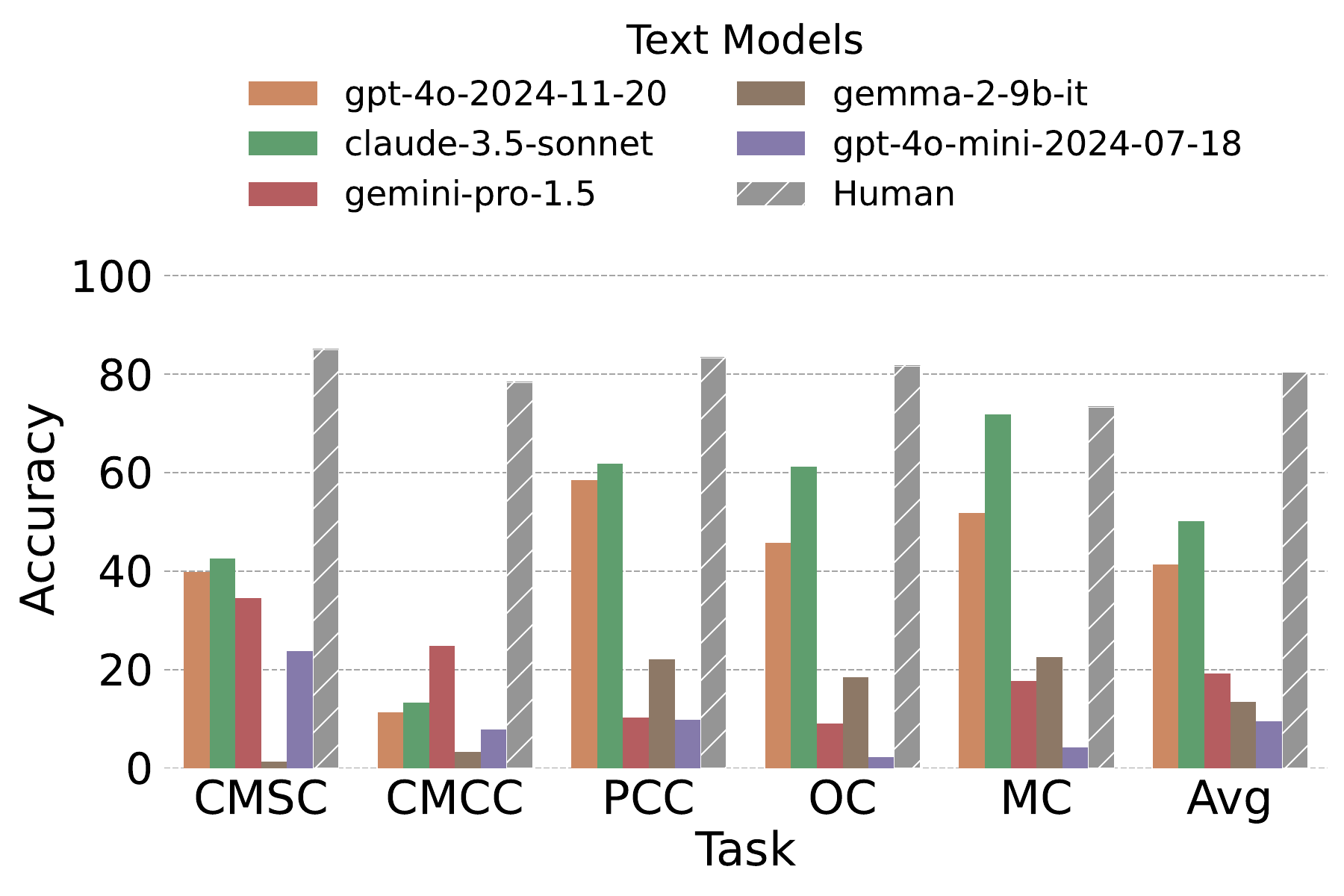}
    \end{subfigure}
    \caption{PBI performance for the top five models for the multimodal (left) and text-only (right) versions of \dataset{}. Human scores are included in both plots for comparison; however, humans were evaluated exclusively in the multimodal setting.}
    \label{fig:top5}
\end{figure*}

\begin{figure*}[t]
    \centering
    \includegraphics[width=\textwidth]{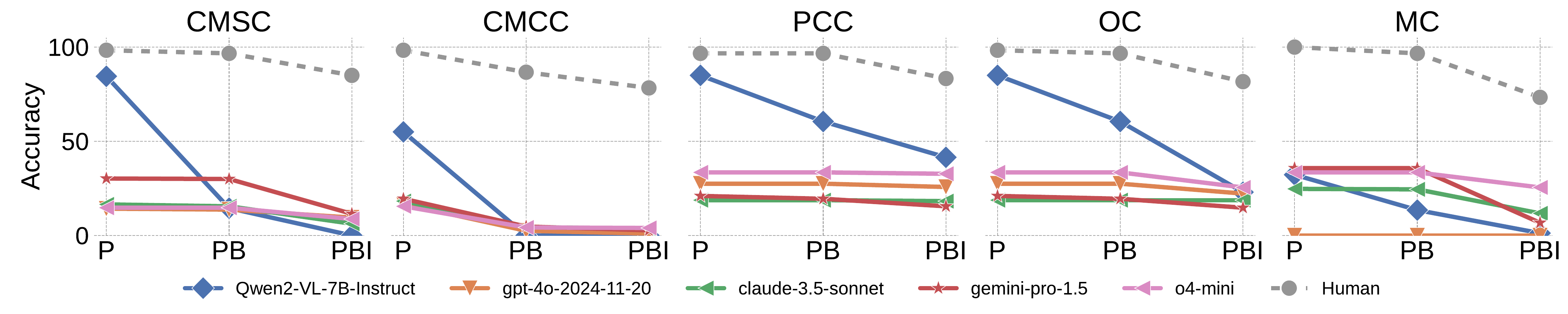}
    \caption{Change in performance across P, PB, and PBI accuracy for both humans and the top five VLMs on \dataset{}. Performance declines from percepts to beliefs to intentions.}
    \label{fig:chain}
\end{figure*}

\section{Experiments}

\subsection{Experimental Setup}

\paragraph{Baseline Models}
We tested 15 baseline models: 
GPT \citep[4o and 4o-mini;][]{gpt4o}, 
o4-mini \cite{o4mini},
Claude~3.5 \citep[Sonnet and Haiku;][]{anthropic-claude}, 
Gemini~1.5 \citep[Pro and Flash;][]{team2023gemini},
Gemini~2.5 Flash \cite{comanici2025gemini},
Llama~3.2~Instruct \citep[1B, 3B, 11B, and 90B;][]{dubey2024llama}, 
Qwen2-VL~Instruct \citep[7B, 72B;][]{wang2024qwen2},
Molmo~7B \citep{deitke2024molmo},
Mistral~7B~Instruct \citep{jiang2023mistral}, 
and Gemma~2~9B Instruct \citep{team2024gemma}.
Language models (LM) were evaluated using the text-only version of the prompts, and vision language models (VLMs) were evaluated both with images and text-only prompts. 
We provide additional details in \ref{app:models}.

\paragraph{Human Study}
We recruited 20 human participants and asked them to answer 45 questions, equally split between tasks.
Participants were shown the same prompt as the models. 
Further details are provided in \ref{app:human-study}.

\paragraph{Metrics} 
For each task, we measured models' accuracy on percept (P), belief (B), and intention (I) questions. 
We then computed two scores: the PB score, which requires correctly answering percept and belief questions ($\text{P} \land \text{B}$), and the PBI score, which requires correctly answering all three question types ($\text{P} \land \text{B} \land \text{I}$).

\begin{figure}[t]
    \centering
    \begin{subfigure}[b]{0.48\textwidth}
        \centering
        \includegraphics[width=\linewidth]{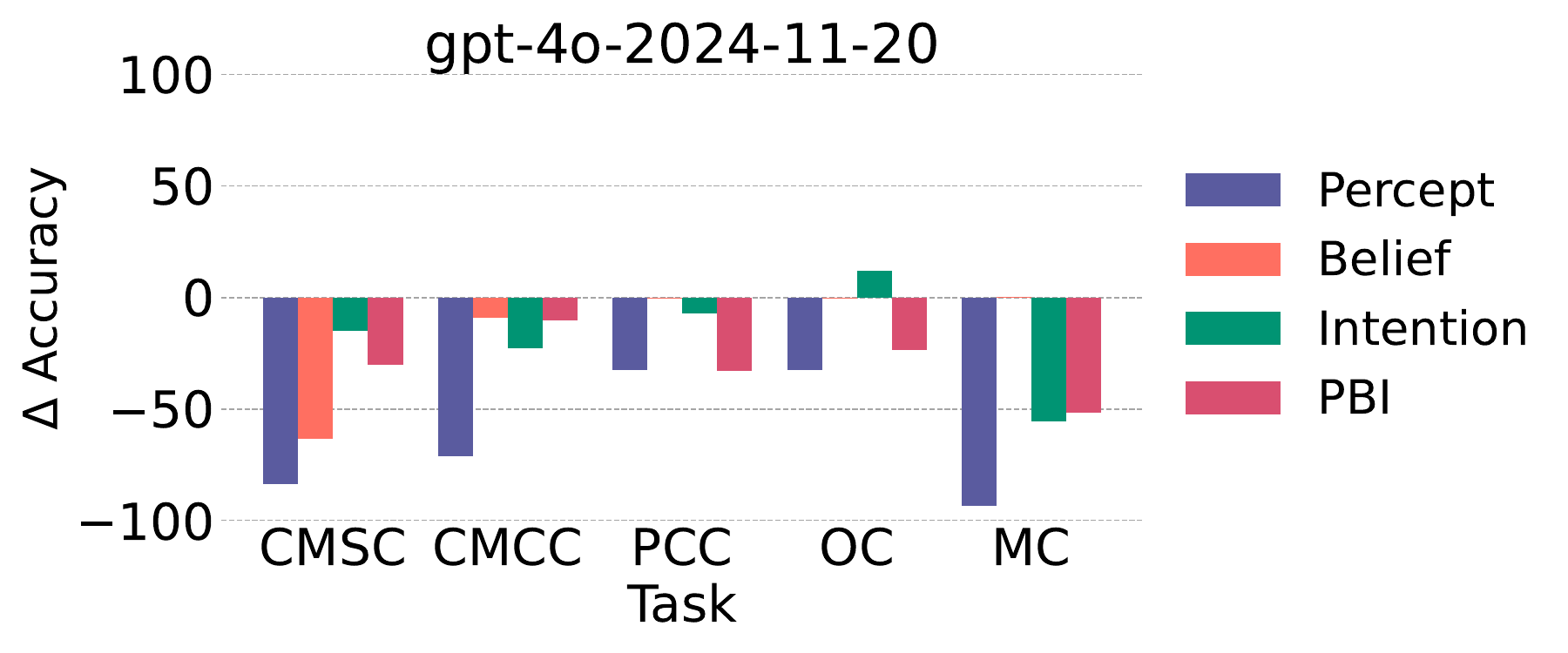}
    \end{subfigure}
    \hfill
    \begin{subfigure}[b]{0.48\textwidth}
        \centering
        \includegraphics[width=\linewidth]{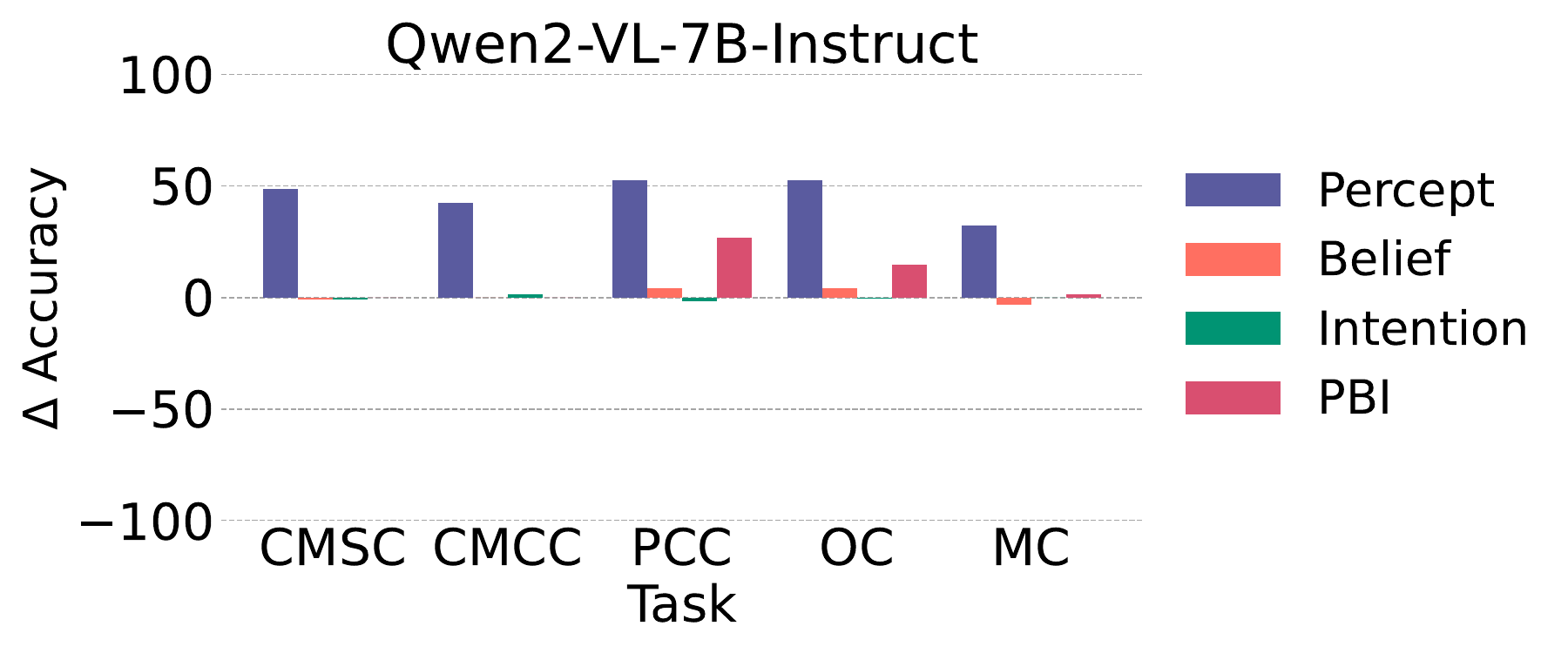}
    \end{subfigure}
    \caption{Difference in accuracy of VLMs when evaluated on the multimodal version of the \dataset{} versus the text-only version.}
    \label{fig:vl-diff}
\end{figure}

\subsection{Results}
\label{sec:results-analyses}

Figure~\ref{fig:top5} reports the PBI accuracy of models on the five tasks in \dataset{}, as well as their average (Avg). 
We focus on the PBI accuracy because it reflects a model's ability to handle the full spectrum of reasoning required to comprehensively solve all tasks in \dataset{}. 
For clarity, the figure includes only the best five models, selected based on their Avg score on the multimodal (Figure~\ref{fig:top5}, left) and text-only versions of \dataset{} (Figure~\ref{fig:top5}, right). 
Detailed results for every combination of task and question type can be found in \ref{app:fullscores}.

Figure~\ref{fig:top5} shows that models' performance significantly lags behind human performance, especially in the multimodal setting. 
While human performance on the tasks ranges from 73\% to 85\%, models generally perform below 30\% (for detailed scores, see Table~\ref{tab:results}). 
No model performs best across all tasks. 
In the multimodal setting, o4-mini is the best model on average (Avg), achieving the highest performance on CMCC, OC, and MC.
Qwen2-7B is the best model on PCC.
PBI scores are higher in the text-only setting, where Claude~3.5~Sonnet performs the best (Figure~\ref{fig:top5}, right). 

\paragraph{P, PB, and PBI Inference}
Accurately representing the complete percept-belief-intention causal graph involves three key steps: inferring the percepts of the target agent, determining the agent’s beliefs based on those percepts, and inferring the agent's intentions based on their beliefs~\citep{jung2024perceptions}. 
Figure~\ref{fig:chain} shows that while models generally perform well in percept inference, their accuracy drops significantly when progressing to PB, and further when moving forward to PBI. 
While humans outperform these models, their accuracy also drops from PB to PBI, albeit less pronounced. 
o4-mini, GPT-4o, Claude~3.5~Sonnet, and Gemini~1.5~Pro are more consistent across all three steps but are limited by lower percept inference accuracy, which also caps their overall performance. 
These findings highlight percept inference as a key foundation for advancing ToM in multimodal models.
Furthermore, our results reveal that even when models achieve reasonable PB inference, they still struggle to transition from PB to PBI.
We provide additional results in \ref{app:pbi-chain}.

\subsection{Error Analysis}
\label{sec:error-analysis}

Figure~\ref{fig:top5} shows that even state-of-the-art models struggle with \dataset{}, indicating a notable gap in their reasoning abilities. 
To gain deeper insights into these shortcomings, we conducted an error analysis by manually inspecting the models' generated outputs.
In particular, we focus on CMCC and MC -- which are the most challenging tasks. 

\paragraph{Modelling Multi-Agent Communication is Challenging}
Compared to CMSC, CMCC is more challenging for models as it involves a communicative event where the target agent is unsure of what has been communicated. 
Despite this, it should be relatively straightforward to infer what has been communicated, given the agents' cooperative intent and their observable initial knowledge (\aone: \bread \blank \tomato \meat, \atwo: \bread \salad \tomato \blank). 
By inspecting the output generated by Llama~3.2~11B\footnote{The best-performing VLM for percept questions on CMCC, see Table~\ref{tab:results}.}, we observe that most errors arise from ignoring that agents can observe each other’s initial knowledge -- although this is made explicit in the prompt.
This oversight leads the model to incorrectly assume that the target agent will believe that if one agent does not explicitly communicate a piece of information, they do not possess it (Example~\ref{box:llama-cmsc-pb-error}).
Next, we examined error cases in the intention questions where both percepts and beliefs were correctly inferred. 
In these instances, we found that errors typically stemmed from incorrect recall of the target agent's knowledge (Example~\ref{box:llama-cmsc-pbi-error}).

\paragraph{Successes and Shortcomings in Modelling Mixed Social Interactions} 
We repeated the previous analysis for the MC task.
While examining the output generated by Llama~3.2~11B\footnote{The best-performing VLM for percept questions on MC, see Table~\ref{tab:results}.}, we found that it accurately considers the agents' attitudes in most cases. 
In successful cases, the model infers that \azero{} prefers not to share information that \atwo{} could learn next (Example~\ref{box:llama-mc-pbi-correct}). 
In failure cases, the model overlooks the fact that if \azero{} communicates a piece of information that both \aone{} and \atwo{} are missing, \aone{} is likely to share it with \atwo{} (Example~\ref{box:llama-mc-pbi-error}).

\subsection{Do VLMs Benefit From Images?} 
\label{sec:l_vs_vl}
We finally compared the performance of VLMs when evaluated on the multimodal version of \dataset{} versus the text-only version. 
Our analysis reveals that different models exhibit different patterns to the inclusion of image inputs. 
For example, GPT-4o does not benefit from the addition of images but performs significantly better on the text-only version of \dataset{} (Figure~\ref{fig:vl-diff}, top). 
This effect is most pronounced for percept questions. 
In contrast, Qwen2-VL~7B shows a clear benefit from the image input, particularly on percept questions (Figure~\ref{fig:vl-diff}, bottom). 
We provide comparisons for all the other evaluated models in \ref{app:l-vs-vl}.
While Claude~3.5~Sonnet and Gemini~Pro~1.5 show similar patterns to GPT-4o, results for other models are mixed, with some benefiting from image inputs on specific tasks and others showing little to no advantage.

\section{Related Work}

Theory of Mind in AI has been studied for more than a decade \citep{baker2011bayesian, eysenbach2016mistaken, rabinowitz2018machine, jara2019theory, liu2023computational, bortoletto2024limits, bortoletto2024explicit, bortoletto2024neural, ruhdorfer2025yokai}. 
Recent advances in LFMs have sparked interest in evaluating their ToM capabilities \citep{achiam2023gpt, ullman2023large}.
While several benchmarks are based on textual variations of the classic Sally-Anne task~\cite{le2019revisiting, sclar2023minding, ma2023tomchallenges, he2023hi, gandhi2024understanding, xu2024opentom, zhou2023far}, others aim to cover a broader range of scenarios \cite{kim2023fantom, chen2024tombench, hou2024entering, gu2024simpletom}, including multimodal settings \cite{bara2021mindcraft, bara2023towards, jin2024mmtom, shi2024muma}. 
Additionally, efforts have been made to enhance LMs' ToM through prompting techniques \citep{zhou2023far, moghaddam2023boosting, wilf2023think} or activation steering \citep{zhu2024language, bortoletto2024benchmarking}.

\section{Discussion and Conclusion}

We introduced \dataset{}, a multimodal benchmark that tests ToM capabilities in environments rich with social interactions and spatial dynamics. 
Featuring up to four agents communicating and moving, it enables the study of cooperative, obstructive, and mixed interactions.
Our evaluations on \dataset{} revealed several important and novel insights. 
First, current \textbf{models perform significantly worse than humans}, both on the multimodal and text-only version of \dataset{} (\S\ref{sec:results-analyses}, Figure~\ref{fig:top5}).
Second, we show that \textbf{models struggle with the critical steps necessary for ToM reasoning} (\S\ref{sec:error-analysis}, Figure~\ref{fig:chain}). 
Notably, even when models perform reasonably well on PB inference, they still struggle to transition to PBI.
Third, our analyses of error cases revealed that \textbf{models are still limited in modelling agent perception, multi-agent communication, and mixed social interactions} (\S\ref{sec:error-analysis}). 
This means they may misinterpret or oversimplify human behaviour in real-world settings, especially in group interactions.
Finally, we found that \textbf{most VLMs do not benefit from visual input} -- highlighting a critical disparity in how models leverage multimodal information to perform ToM tasks (\S\ref{sec:l_vs_vl}, Figure~\ref{fig:vl-diff}).
This suggests a gap in how models understand and integrate context, which is vital for interpreting visual cues during social interactions.

\section*{Limitations}

One limitation of \dataset{} lies in its synthetic grid world environment, which is simpler than the real world. 
However, this simplicity does not impair the core abilities that \dataset{} targets -- reasoning about agents' mental states in spatially grounded interactions. 
There is a clear advantage that makes synthetic benchmarks well suited for studying ToM, at least at the current state of research: real-world tasks often require common sense reasoning skills that models do not fully have yet, and that will function as a confounder making models' performance on ToM inference hard to judge (see \citet{gandhi2024understanding}'s discussion about \citep{shapira2024clever}). 
Synthetic benchmarks like ToM-SSI allow us to reduce these factors and to design tasks that better focus on evaluating core ToM abilities.
Our setup also makes the environment less prone to hallucinations, which can hinder the performance of current vision-language models in complex simulations~\citep{jin2024mmtom}, while also making hallucinations easier to identify if they occur.
In addition, using images instead of videos avoids sampling issues or additional processing steps that might exclude important video frames. 

A second limitation is that, while our work covers a broader set of interactions compared to previous work, it still does not cover all possible social scenarios. 
Future work could consider, for example, \textit{exploitative interactions}, where an agent uses the other's resources or efforts for their own gain.
Moreover, while \dataset{} presents scenarios about a single group of interacting agents, future work could extend it to multiple \textit{social groups} with agents in the same social group sharing common goals. 
This can be achieved by extending \dataset{}'s generation pipeline with suitable templates.

Finally, while studying inference-time methods to improve performance goes beyond the scope of our current work,  exploring the effects of CoT \citep{wei2022chain} or other methods (e.g. SimToM \cite{wilf2023think}, TimeToM \cite{hou2024timetom}, or PercepToM \citep{jung2024perceptions}) is an interesting research direction for future work.

\section*{Acknowledgements}
The authors thank the International Max Planck Research School for Intelligent Systems (IMPRS-IS) for supporting C. Ruhdorfer.

\bibliography{custom,tom_ref}

\clearpage
\appendix
\section{Appendix}
\label{sec:appendix}

\subsection{Societal Impact}
While our work is foundational and remains distant from specific applications with direct societal impact, it’s important to recognise the ethical implications of modelling and predicting mental states. 
Handling sensitive aspects of individuals' inner experiences requires careful consideration to avoid reinforcing biases or misunderstanding psychological nuances.

\subsection{Gridworld}

\begin{table*}[t]
\centering
\resizebox{\linewidth}{!}{
\begin{tabular}{lccccccccccccccccccccc}
\toprule
\textbf{Model} & \textbf{Vision} & \multicolumn{4}{c}{\textbf{CMSC}} & \multicolumn{4}{c}{\textbf{CMCC}} & \multicolumn{4}{c}{\textbf{PCC}} & \multicolumn{4}{c}{\textbf{OC}} & \multicolumn{4}{c}{\textbf{MC}} \\
\cline{3-22}
& & P & B & I & PBI & P & B & I & PBI & P & B & I & PBI & P & B & I & PBI & P & B & I & PBI \\
\midrule
\texttt{Human} & $\checkmark$ & $98.3$ & $98.3$ & $85.0$ & \cellcolor{gray!20}$85.0$ & $98.3$ & $88.3$ & $85.0$ & \cellcolor{gray!20}$78.3$ & $96.7$ & $98.3$ & $86.7$ & \cellcolor{gray!20}$83.3$ & $98.3$ & $98.3$ & $81.7$ & \cellcolor{gray!20}$81.7$ & $100.0$ & $96.7$ & $73.3$ & \cellcolor{gray!20}$73.3$ \\
\texttt{Llama-3.2-11B-Vision-Instruct} &  & $82.5$ & $22.5$ & $13.5$ & \cellcolor{gray!20}$2.2$ & $64.2$ & $0.0$ & $17.2$ & \cellcolor{gray!20}$0.0$ & $55.5$ & $60.5$ & $59.5$ & \cellcolor{gray!20}$20.2$ & $55.5$ & $60.5$ & $33.5$ & \cellcolor{gray!20}$10.2$ & $65.8$ & $35.5$ & $22.0$ & \cellcolor{gray!20}$5.2$ \\
\texttt{Llama-3.2-11B-Vision-Instruct} & $\checkmark$ & $87.8$ & $25.2$ & $5.2$ & \cellcolor{gray!20}$1.0$ & $65.2$ & $1.0$ & $10.8$ & \cellcolor{gray!20}$0.0$ & $73.8$ & $56.2$ & $67.8$ & \cellcolor{gray!20}$27.3$ & $73.8$ & $56.2$ & $21.5$ & \cellcolor{gray!20}$8.0$ & $61.5$ & $40.2$ & $20.2$ & \cellcolor{gray!20}$5.0$ \\
\texttt{Llama-3.2-1B-Instruct} &  & $50.2$ & $23.8$ & $30.2$ & \cellcolor{gray!20}$3.0$ & $48.2$ & $24.5$ & $28.5$ & \cellcolor{gray!20}$4.5$ & $51.0$ & $36.2$ & $51.0$ & \cellcolor{gray!20}$11.2$ & $51.0$ & $36.2$ & $27.0$ & \cellcolor{gray!20}$6.2$ & $53.8$ & $22.2$ & $23.2$ & \cellcolor{gray!20}$3.0$ \\
\texttt{Llama-3.2-3B-Instruct} &  & $63.7$ & $18.2$ & $20.8$ & \cellcolor{gray!20}$2.5$ & $65.2$ & $0.0$ & $18.0$ & \cellcolor{gray!20}$0.0$ & $63.5$ & $63.0$ & $80.2$ & \cellcolor{gray!20}$31.8$ & $63.5$ & $63.0$ & $20.0$ & \cellcolor{gray!20}$8.8$ & $46.2$ & $30.0$ & $22.8$ & \cellcolor{gray!20}$3.5$ \\
\texttt{Mistral-7B-Instruct-v0.3} &  & $72.8$ & $24.5$ & $16.0$ & \cellcolor{gray!20}$4.0$ & $46.2$ & $0.0$ & $15.8$ & \cellcolor{gray!20}$0.0$ & $60.5$ & $61.3$ & $66.8$ & \cellcolor{gray!20}$22.2$ & $60.5$ & $61.3$ & $26.0$ & \cellcolor{gray!20}$8.2$ & $25.0$ & $46.5$ & $22.2$ & \cellcolor{gray!20}$2.2$ \\
\texttt{Molmo-7B-D-0924} &  & $56.0$ & $23.8$ & $38.8$ & \cellcolor{gray!20}$5.2$ & $59.2$ & $4.8$ & $34.2$ & \cellcolor{gray!20}$1.0$ & $44.5$ & $44.0$ & $37.5$ & \cellcolor{gray!20}$7.2$ & $44.5$ & $44.0$ & $17.8$ & \cellcolor{gray!20}$2.8$ & $51.0$ & $36.8$ & $10.2$ & \cellcolor{gray!20}$1.2$ \\
\texttt{Molmo-7B-D-0924} & $\checkmark$ & $43.5$ & $29.8$ & $33.0$ & \cellcolor{gray!20}$5.0$ & $51.0$ & $4.0$ & $28.0$ & \cellcolor{gray!20}$0.2$ & $41.5$ & $44.0$ & $41.0$ & \cellcolor{gray!20}$6.8$ & $41.5$ & $44.0$ & $16.0$ & \cellcolor{gray!20}$3.8$ & $51.0$ & $39.0$ & $14.8$ & \cellcolor{gray!20}$2.2$ \\
\texttt{Qwen2-VL-7B-Instruct} &  & $35.8$ & $17.0$ & $1.8$ & \cellcolor{gray!20}$0.0$ & $12.8$ & $0.0$ & $4.2$ & \cellcolor{gray!20}$0.0$ & $32.2$ & $66.8$ & $67.2$ & \cellcolor{gray!20}$14.5$ & $32.2$ & $66.8$ & $39.2$ & \cellcolor{gray!20}$8.2$ & $0.0$ & $42.2$ & $10.2$ & \cellcolor{gray!20}$0.0$ \\
\texttt{Qwen2-VL-7B-Instruct} & $\checkmark$ & $84.5$ & $16.0$ & $0.8$ & \cellcolor{gray!20}$0.0$ & $55.0$ & $0.0$ & $5.5$ & \cellcolor{gray!20}$0.0$ & $85.0$ & $71.0$ & $65.8$ & \cellcolor{gray!20}$41.5$ & $85.0$ & $71.0$ & $38.8$ & \cellcolor{gray!20}$23.0$ & $32.2$ & $39.0$ & $10.2$ & \cellcolor{gray!20}$1.2$ \\
\texttt{claude-3.5-haiku-20241022} &  & $68.8$ & $84.2$ & $26.0$ & \cellcolor{gray!20}$16.5$ & $69.2$ & $3.2$ & $76.2$ & \cellcolor{gray!20}$2.5$ & $11.2$ & $92.2$ & $76.5$ & \cellcolor{gray!20}$7.8$ & $11.2$ & $92.2$ & $23.5$ & \cellcolor{gray!20}$2.0$ & $72.8$ & $100.0$ & $4.2$ & \cellcolor{gray!20}$3.5$ \\
\texttt{claude-3.5-haiku-20241022} & $\checkmark$ & $5.2$ & $75.8$ & $23.2$ & \cellcolor{gray!20}$0.5$ & $12.0$ & $3.0$ & $70.0$ & \cellcolor{gray!20}$0.0$ & $4.8$ & $91.8$ & $81.2$ & \cellcolor{gray!20}$4.2$ & $4.8$ & $91.8$ & $17.5$ & \cellcolor{gray!20}$0.2$ & $46.5$ & $100.0$ & $3.0$ & \cellcolor{gray!20}$1.5$ \\
\texttt{claude-3.5-sonnet} &  & $85.2$ & $100.0$ & $49.0$ & \cellcolor{gray!20}$42.5$ & $75.2$ & $21.2$ & $75.5$ & \cellcolor{gray!20}$13.2$ & $64.5$ & $99.8$ & $95.5$ & \cellcolor{gray!20}$61.8$ & $64.5$ & $99.8$ & $95.2$ & \cellcolor{gray!20}$61.3$ & $84.5$ & $100.0$ & $85.2$ & \cellcolor{gray!20}$71.8$ \\
\texttt{claude-3.5-sonnet} & $\checkmark$ & $16.5$ & $91.0$ & $20.5$ & \cellcolor{gray!20}$6.2$ & $18.2$ & $13.0$ & $75.2$ & \cellcolor{gray!20}$3.5$ & $18.8$ & $99.8$ & $97.8$ & \cellcolor{gray!20}$18.2$ & $18.8$ & $99.8$ & $97.8$ & \cellcolor{gray!20}$18.8$ & $24.8$ & $98.8$ & $49.8$ & \cellcolor{gray!20}$11.8$ \\
\texttt{gemini-2.5-flash} &  & $8.2$ & $7.2$ & $0.0$ & \cellcolor{gray!20}$0.0$ & $10.0$ & $0.2$ & $2.0$ & \cellcolor{gray!20}$0.0$ & $14.8$ & $4.0$ & $0.2$ & \cellcolor{gray!20}$0.0$ & $14.8$ & $4.0$ & $0.2$ & \cellcolor{gray!20}$0.0$ & $13.0$ & $2.0$ & $0.2$ & \cellcolor{gray!20}$0.0$ \\
\texttt{gemini-2.5-flash} & $\checkmark$ & $6.0$ & $10.0$ & $1.8$ & \cellcolor{gray!20}$0.2$ & $6.8$ & $2.8$ & $6.2$ & \cellcolor{gray!20}$0.0$ & $20.8$ & $24.0$ & $14.5$ & \cellcolor{gray!20}$1.2$ & $20.8$ & $24.0$ & $4.2$ & \cellcolor{gray!20}$0.2$ & $12.2$ & $12.8$ & $6.5$ & \cellcolor{gray!20}$0.0$ \\
\texttt{gemini-flash-1.5} &  & $45.0$ & $100.0$ & $16.5$ & \cellcolor{gray!20}$7.5$ & $15.2$ & $1.8$ & $57.5$ & \cellcolor{gray!20}$0.0$ & $15.5$ & $78.8$ & $35.8$ & \cellcolor{gray!20}$4.5$ & $15.5$ & $78.8$ & $29.5$ & \cellcolor{gray!20}$4.2$ & $41.5$ & $99.8$ & $6.2$ & \cellcolor{gray!20}$2.8$ \\
\texttt{gemini-flash-1.5} & $\checkmark$ & $21.0$ & $93.2$ & $26.0$ & \cellcolor{gray!20}$5.5$ & $12.2$ & $1.8$ & $46.8$ & \cellcolor{gray!20}$0.0$ & $30.0$ & $84.2$ & $43.2$ & \cellcolor{gray!20}$8.8$ & $30.0$ & $84.2$ & $39.5$ & \cellcolor{gray!20}$9.8$ & $29.2$ & $100.0$ & $10.0$ & \cellcolor{gray!20}$3.5$ \\
\texttt{gemini-pro-1.5} &  & $71.8$ & $100.0$ & $48.5$ & \cellcolor{gray!20}$34.5$ & $44.2$ & $75.8$ & $70.0$ & \cellcolor{gray!20}$24.8$ & $12.8$ & $92.5$ & $86.5$ & \cellcolor{gray!20}$10.2$ & $12.8$ & $92.5$ & $76.2$ & \cellcolor{gray!20}$9.0$ & $54.5$ & $100.0$ & $30.8$ & \cellcolor{gray!20}$17.8$ \\
\texttt{gemini-pro-1.5} & $\checkmark$ & $30.2$ & $99.2$ & $29.2$ & \cellcolor{gray!20}$11.5$ & $19.5$ & $21.2$ & $55.5$ & \cellcolor{gray!20}$2.8$ & $21.0$ & $90.2$ & $88.5$ & \cellcolor{gray!20}$15.5$ & $21.0$ & $90.2$ & $78.5$ & \cellcolor{gray!20}$14.8$ & $35.8$ & $100.0$ & $18.0$ & \cellcolor{gray!20}$6.8$ \\
\texttt{gemma-2-9b-it} &  & $99.0$ & $81.2$ & $1.5$ & \cellcolor{gray!20}$1.2$ & $38.8$ & $32.5$ & $24.5$ & \cellcolor{gray!20}$3.2$ & $69.2$ & $74.8$ & $42.0$ & \cellcolor{gray!20}$22.0$ & $69.2$ & $74.8$ & $39.0$ & \cellcolor{gray!20}$18.5$ & $99.2$ & $96.5$ & $24.2$ & \cellcolor{gray!20}$22.5$ \\
\texttt{gpt-4o-2024-11-20} &  & $98.0$ & $98.8$ & $41.2$ & \cellcolor{gray!20}$39.8$ & $88.0$ & $19.8$ & $54.8$ & \cellcolor{gray!20}$11.2$ & $60.0$ & $99.8$ & $97.2$ & \cellcolor{gray!20}$58.5$ & $60.0$ & $99.8$ & $74.2$ & \cellcolor{gray!20}$45.8$ & $93.2$ & $99.8$ & $55.5$ & \cellcolor{gray!20}$51.7$ \\
\texttt{gpt-4o-2024-11-20} & $\checkmark$ & $14.2$ & $35.5$ & $26.2$ & \cellcolor{gray!20}$9.5$ & $16.8$ & $10.5$ & $32.0$ & \cellcolor{gray!20}$1.0$ & $27.5$ & $99.2$ & $90.2$ & \cellcolor{gray!20}$25.8$ & $27.5$ & $99.2$ & $86.2$ & \cellcolor{gray!20}$22.2$ & $0.0$ & $100.0$ & $0.0$ & \cellcolor{gray!20}$0.0$ \\
\texttt{gpt-4o-mini-2024-07-18} &  & $65.5$ & $100.0$ & $35.8$ & \cellcolor{gray!20}$23.8$ & $58.8$ & $25.8$ & $47.5$ & \cellcolor{gray!20}$7.8$ & $16.0$ & $86.2$ & $57.2$ & \cellcolor{gray!20}$9.8$ & $16.0$ & $86.2$ & $10.0$ & \cellcolor{gray!20}$2.2$ & $60.0$ & $99.8$ & $6.8$ & \cellcolor{gray!20}$4.2$ \\
\texttt{gpt-4o-mini-2024-07-18} & $\checkmark$ & $23.0$ & $98.8$ & $34.0$ & \cellcolor{gray!20}$7.8$ & $33.8$ & $9.0$ & $59.8$ & \cellcolor{gray!20}$1.8$ & $16.8$ & $86.5$ & $64.0$ & \cellcolor{gray!20}$10.8$ & $16.8$ & $86.5$ & $11.2$ & \cellcolor{gray!20}$2.0$ & $25.2$ & $100.0$ & $6.8$ & \cellcolor{gray!20}$2.0$ \\
\texttt{llama-3.2-90b-vision-instruct} &  & $30.2$ & $75.8$ & $26.2$ & \cellcolor{gray!20}$6.2$ & $6.5$ & $0.0$ & $43.2$ & \cellcolor{gray!20}$0.0$ & $8.2$ & $91.2$ & $66.0$ & \cellcolor{gray!20}$4.2$ & $8.2$ & $91.2$ & $16.5$ & \cellcolor{gray!20}$2.2$ & $52.2$ & $100.0$ & $3.2$ & \cellcolor{gray!20}$1.0$ \\
\texttt{llama-3.2-90b-vision-instruct} & $\checkmark$ & $9.2$ & $55.8$ & $23.0$ & \cellcolor{gray!20}$0.8$ & $1.8$ & $4.8$ & $78.8$ & \cellcolor{gray!20}$0.0$ & $14.0$ & $72.5$ & $82.8$ & \cellcolor{gray!20}$8.5$ & $14.0$ & $72.5$ & $42.5$ & \cellcolor{gray!20}$3.5$ & $25.8$ & $66.8$ & $17.8$ & \cellcolor{gray!20}$3.2$ \\
\texttt{o4-mini-2025-04-16} & $\checkmark$ & $14.8$ & $69.0$ & $25.2$ & \cellcolor{gray!20}$8.8$ & $15.5$ & $17.0$ & $34.2$ & \cellcolor{gray!20}$4.0$ & $33.5$ & $100.0$ & $95.5$ & \cellcolor{gray!20}$32.8$ & $33.5$ & $100.0$ & $86.8$ & \cellcolor{gray!20}$25.5$ & $33.5$ & $100.0$ & $74.2$ & \cellcolor{gray!20}$25.5$ \\
\texttt{qwen-2-vl-72b-instruct} &  & $59.2$ & $91.0$ & $3.2$ & \cellcolor{gray!20}$2.5$ & $7.8$ & $0.2$ & $21.2$ & \cellcolor{gray!20}$0.0$ & $50.0$ & $76.2$ & $52.2$ & \cellcolor{gray!20}$17.8$ & $50.0$ & $76.2$ & $50.5$ & \cellcolor{gray!20}$18.8$ & $36.2$ & $99.5$ & $8.5$ & \cellcolor{gray!20}$2.2$ \\
\texttt{qwen-2-vl-72b-instruct} & $\checkmark$ & $21.2$ & $89.5$ & $7.8$ & \cellcolor{gray!20}$1.8$ & $0.8$ & $6.5$ & $26.8$ & \cellcolor{gray!20}$0.0$ & $49.5$ & $78.5$ & $50.7$ & \cellcolor{gray!20}$17.0$ & $49.5$ & $78.5$ & $51.7$ & \cellcolor{gray!20}$21.8$ & $13.2$ & $99.8$ & $7.5$ & \cellcolor{gray!20}$1.0$ \\
\bottomrule
\end{tabular}
}
\caption{Models' accuracy across the three question types (P: Perception, B: Belief, I: Intent) for each task in \dataset{}.}
\label{tab:results}
\end{table*}

\subsubsection{Gridworld Representations}
\begin{figure}[t]
    \centering
    \begin{minipage}[b]{0.48\textwidth}
        \centering
        \includegraphics[width=\linewidth]{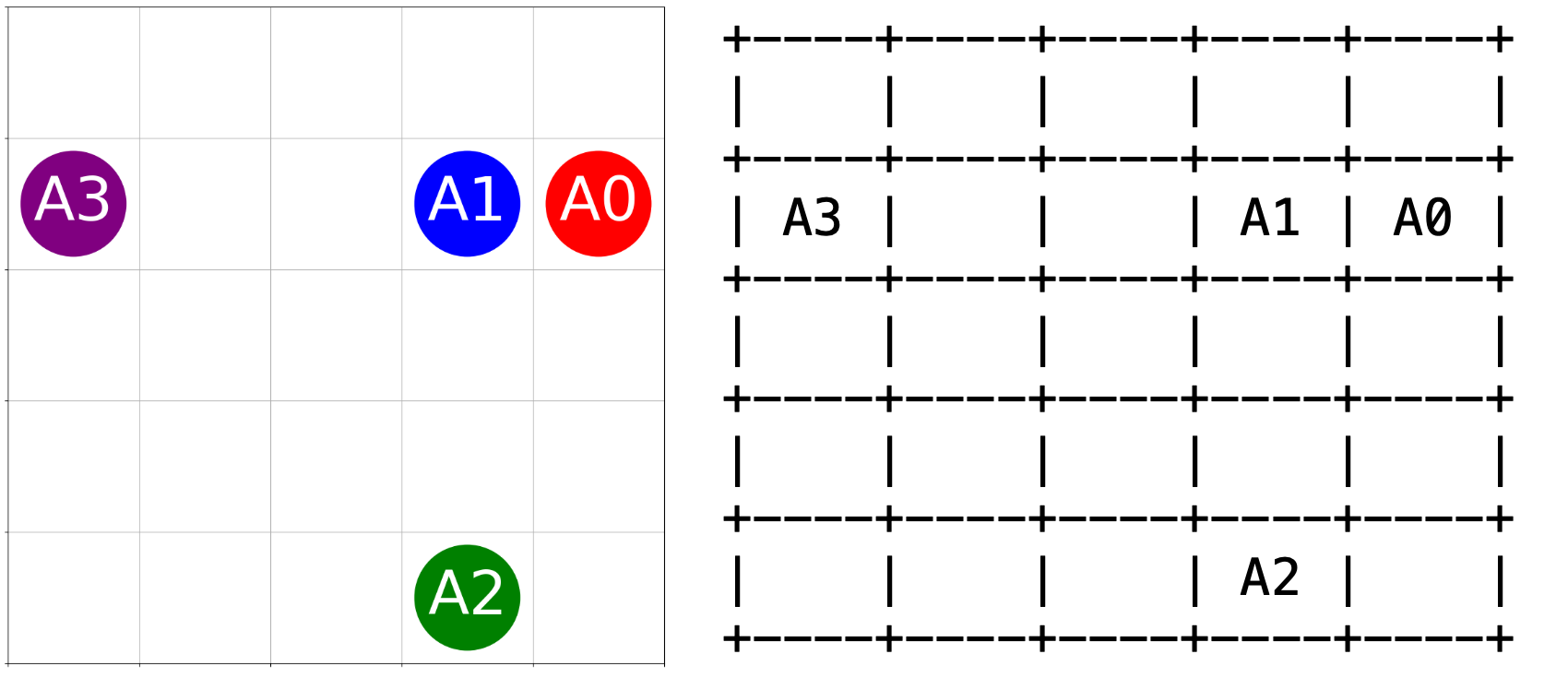}
        \caption{By using grid world environments as a ground for social interactions, \dataset{} allows us to evaluate large language and vision-and-language models by providing equivalent grid representations.}
        \label{fig:grid}
    \end{minipage}
    \hfill
    \begin{minipage}[b]{0.48\textwidth}
        \centering
        \includegraphics[width=\linewidth]{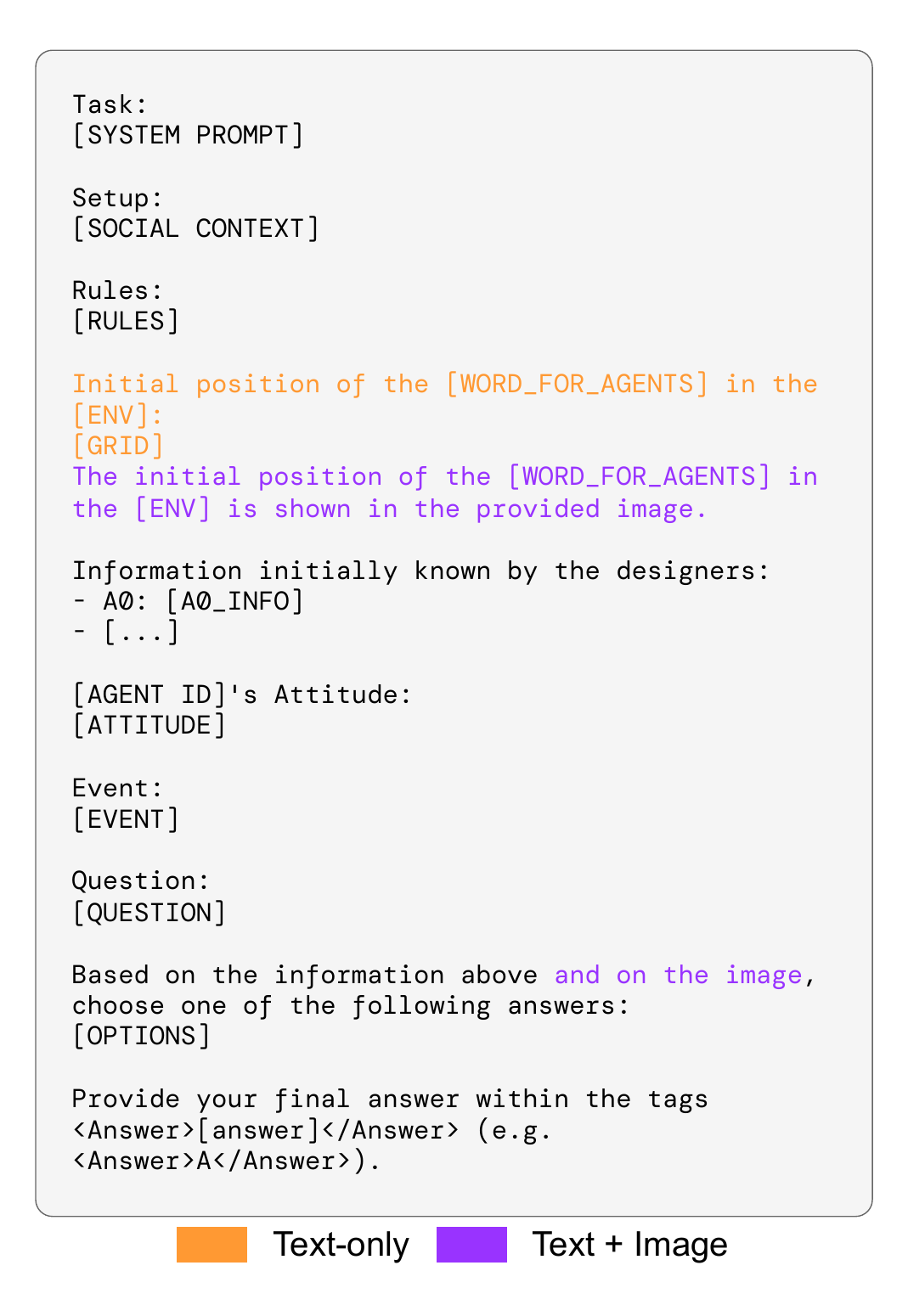}
        \caption{Structure of prompts in \dataset{}.}
        \label{fig:prompt-structure}
    \end{minipage}
\end{figure}

Figure~\ref{fig:grid} represents the two different representation of the grid included in the standard, multimodal version of \dataset{} and in the text-only version. 

\begin{figure*}[t]
    \centering
    \begin{subfigure}[b]{\textwidth}
        \centering
        \includegraphics[width=\linewidth]{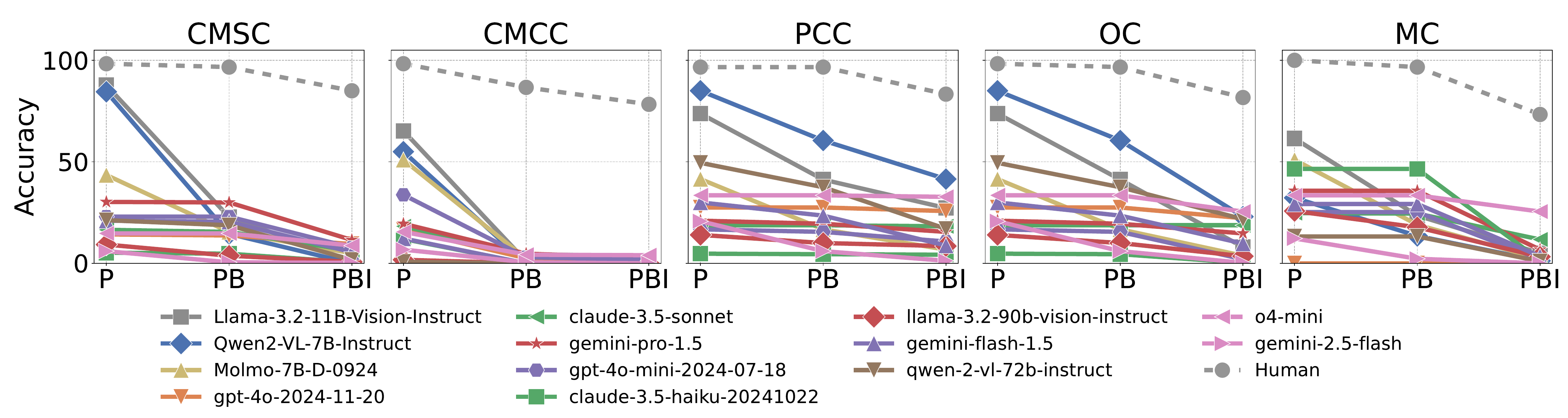}
        \caption{VLMs evaluated on \dataset{}.}
    \end{subfigure}
    \hfill
    \begin{subfigure}[b]{\textwidth}
        \centering
        \includegraphics[width=\linewidth]{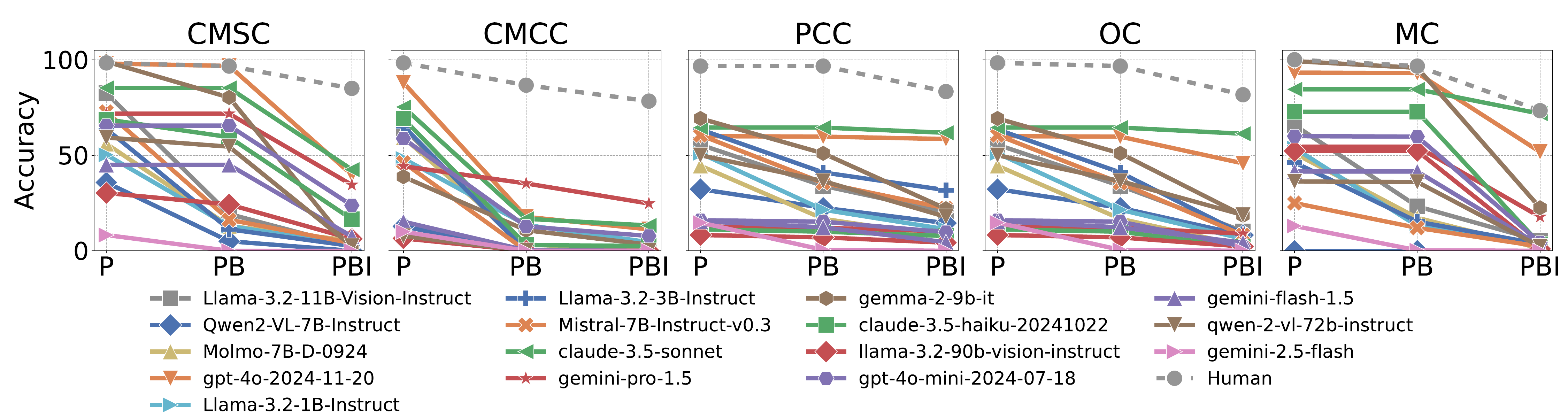}
        \caption{LMs evaluated on the text-only version of \dataset{}.}
    \end{subfigure}
    \caption{Change in performance across P, PB, and PBI accuracy for both humans and models on \dataset{}.}
    \label{fig:chain-all}
\end{figure*}

\begin{figure*}[t]
    \centering
    \begin{subfigure}[b]{0.48\textwidth}
        \centering
        \includegraphics[width=\linewidth]{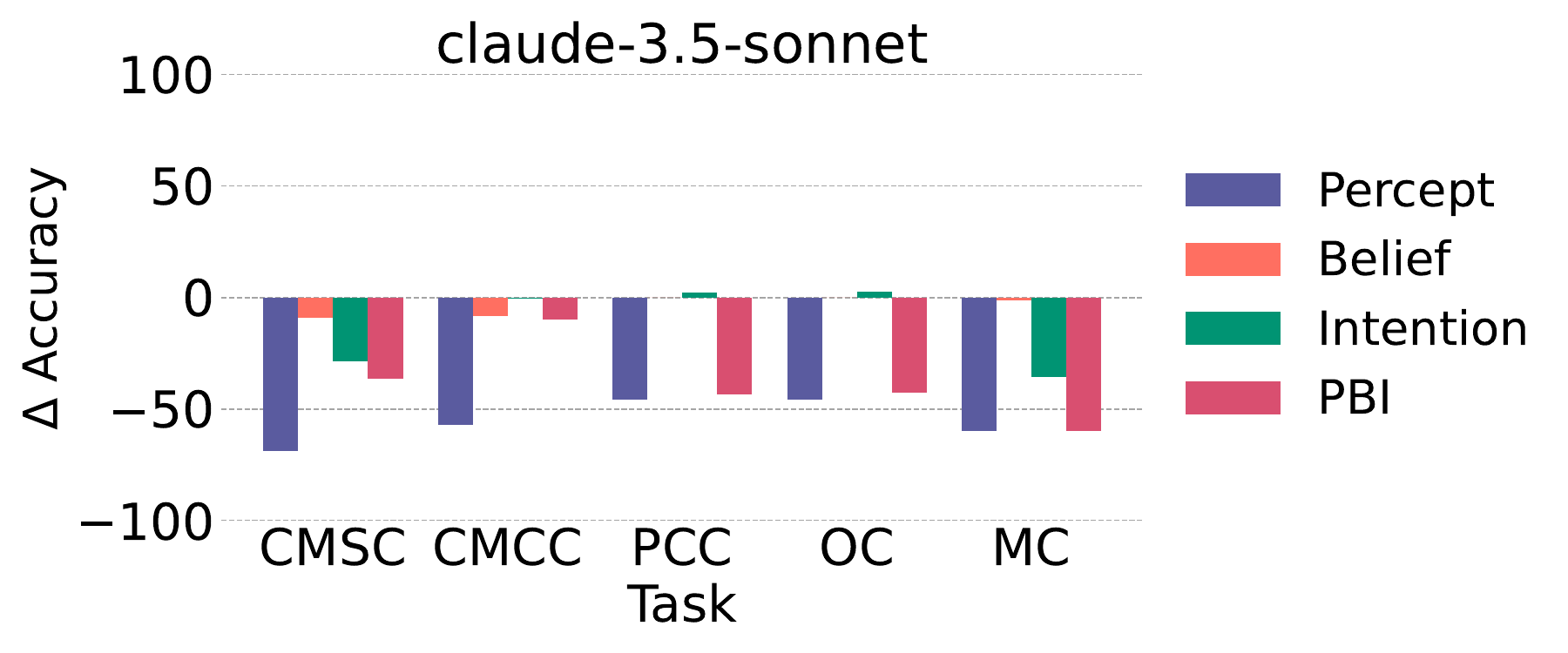}
    \end{subfigure}
    \hfill
    \begin{subfigure}[b]{0.48\textwidth}
        \centering
        \includegraphics[width=\linewidth]{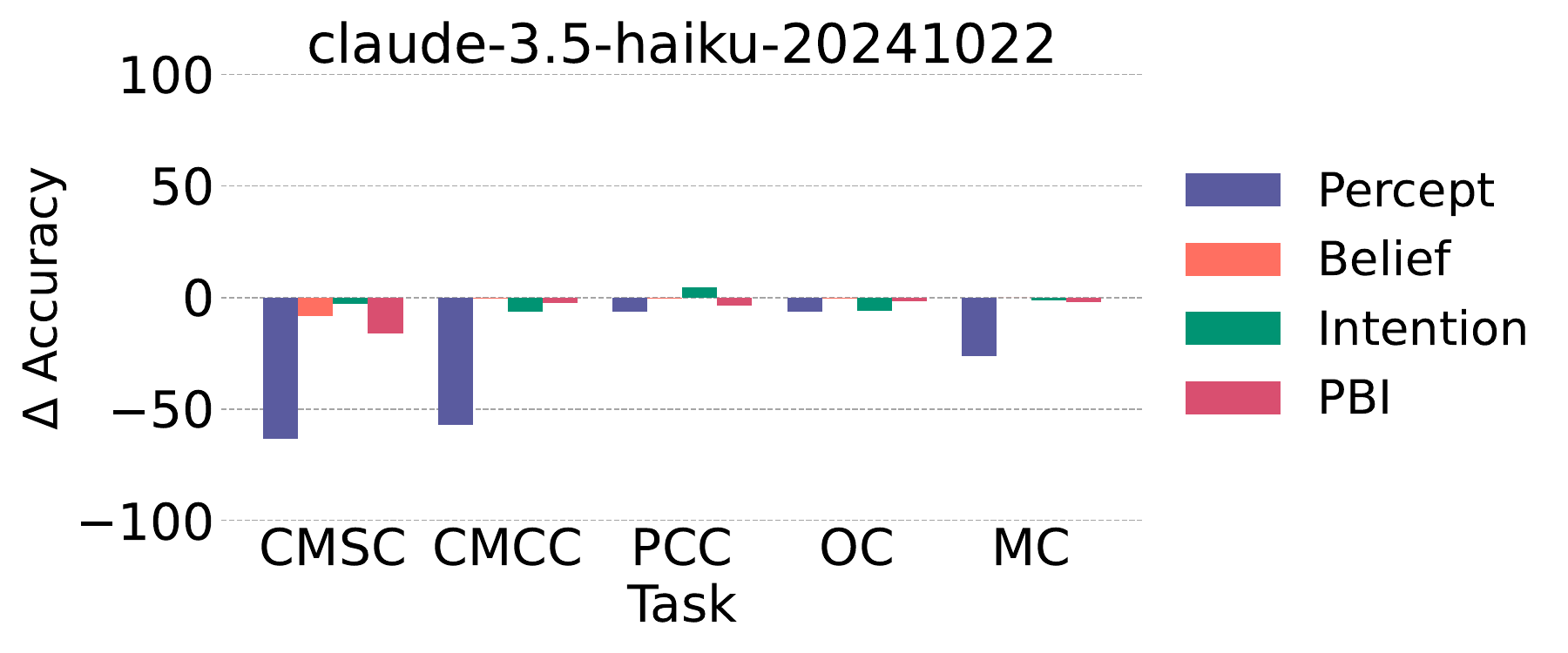}
    \end{subfigure}
    \hfill
    \begin{subfigure}[b]{0.48\textwidth}
        \centering
        \includegraphics[width=\linewidth]{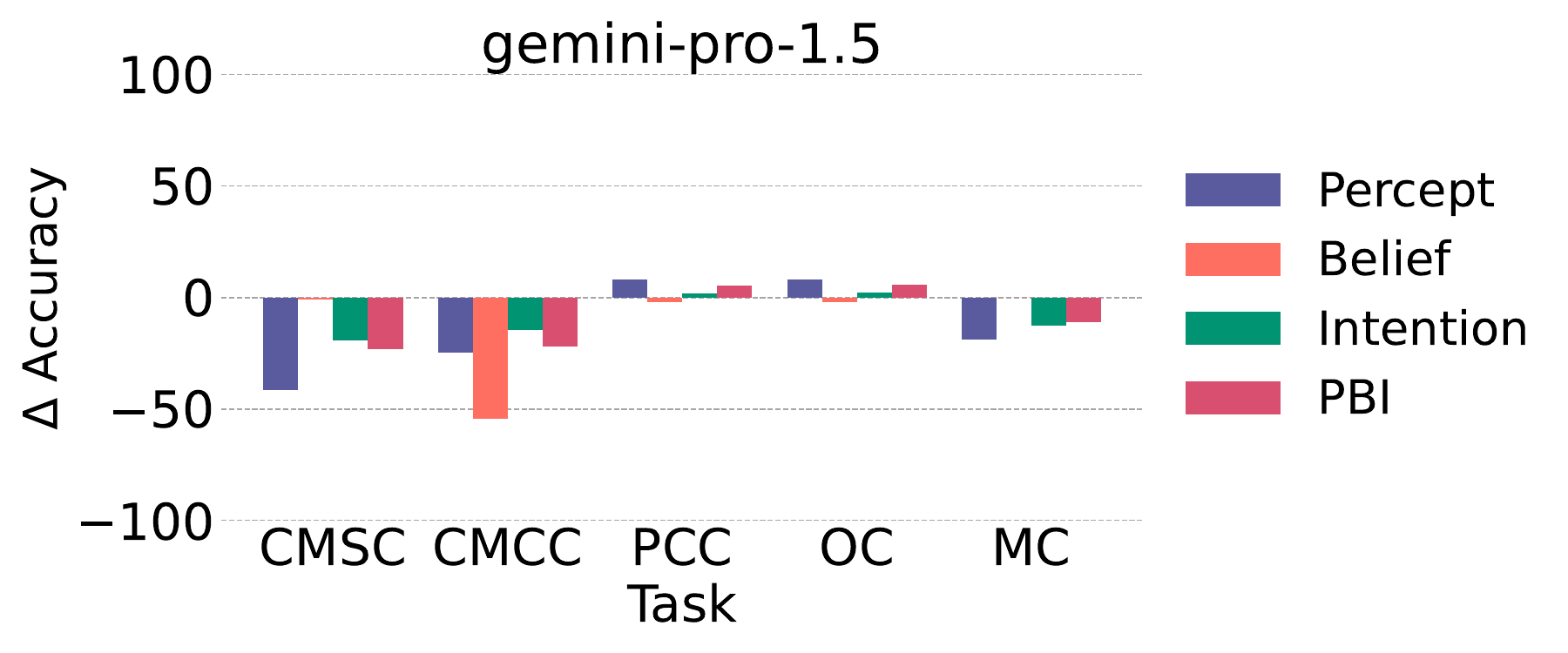}
    \end{subfigure}
    \hfill
    \begin{subfigure}[b]{0.48\textwidth}
        \centering
        \includegraphics[width=\linewidth]{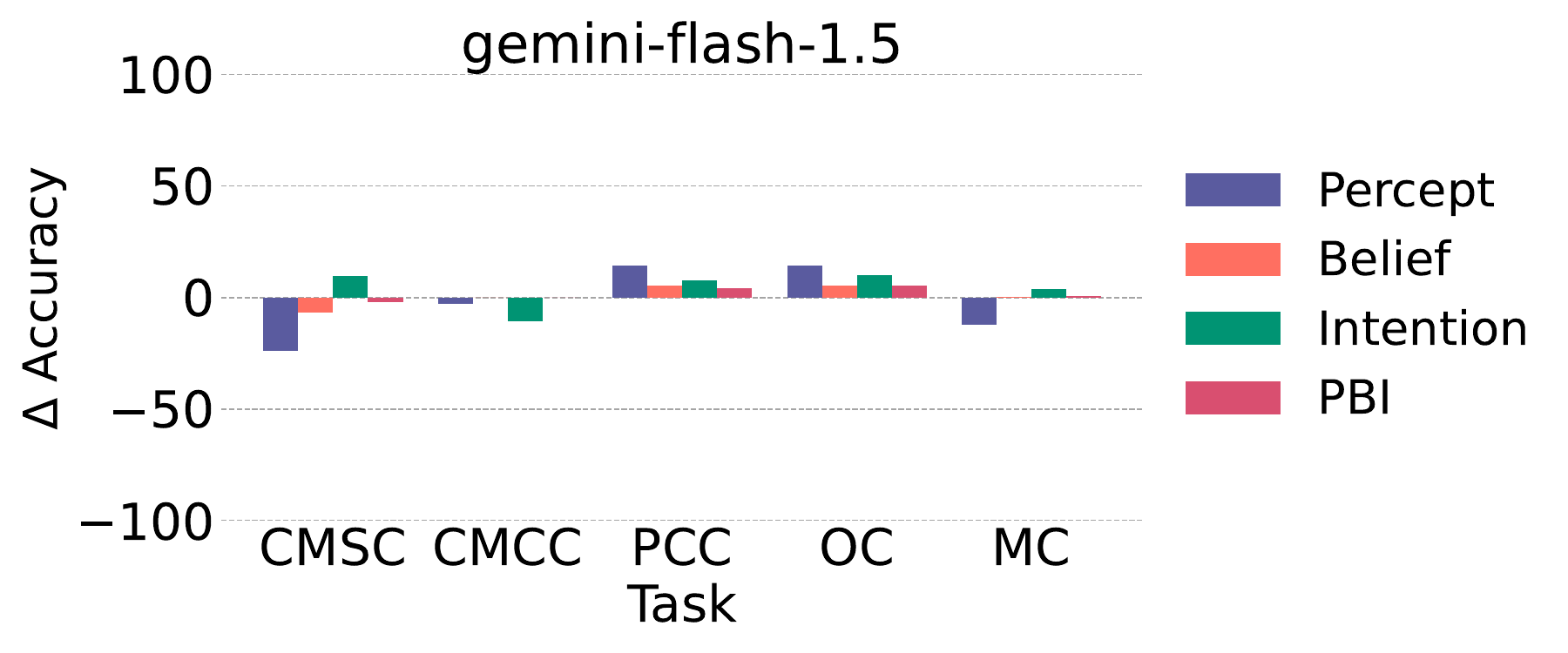}
    \end{subfigure}
    \hfill
    \begin{subfigure}[b]{0.48\textwidth}
        \centering
        \includegraphics[width=\linewidth]{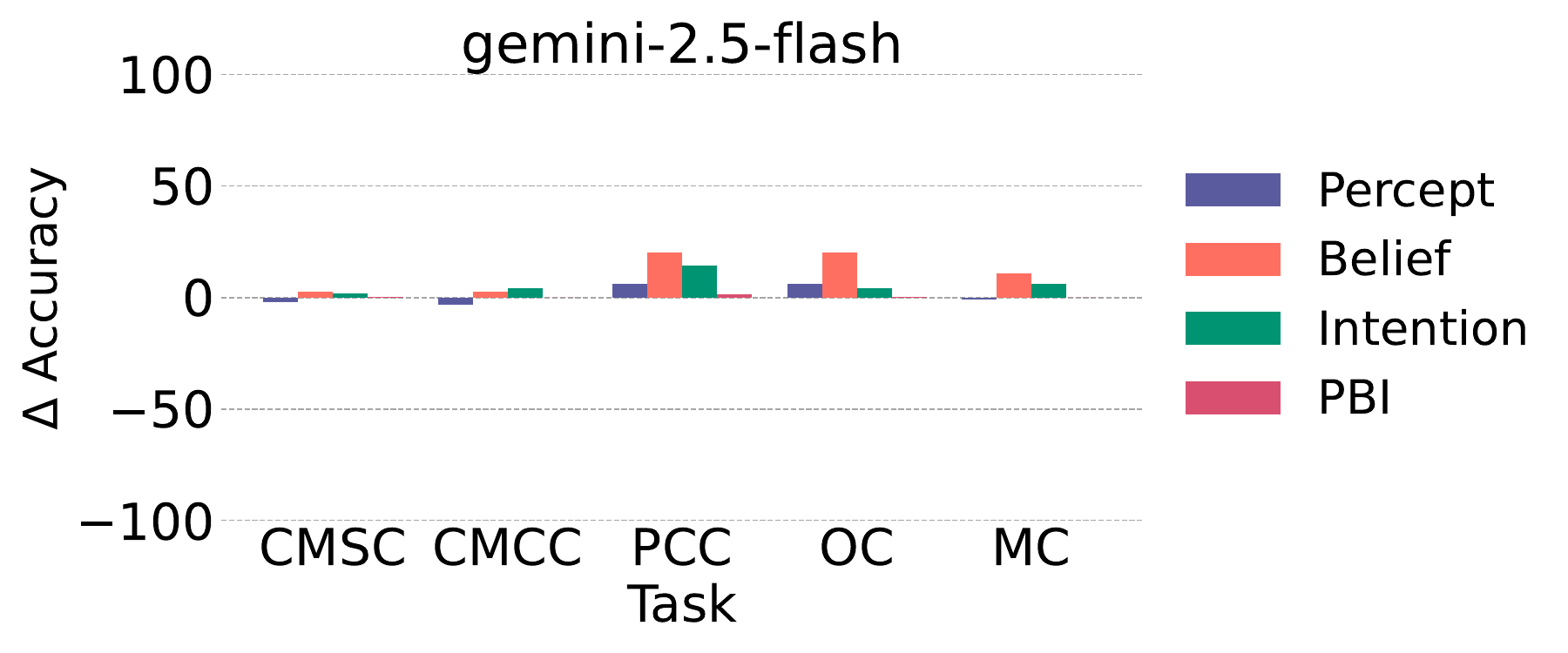}
    \end{subfigure}
    \hfill
    \begin{subfigure}[b]{0.48\textwidth}
        \centering
        \includegraphics[width=\linewidth]{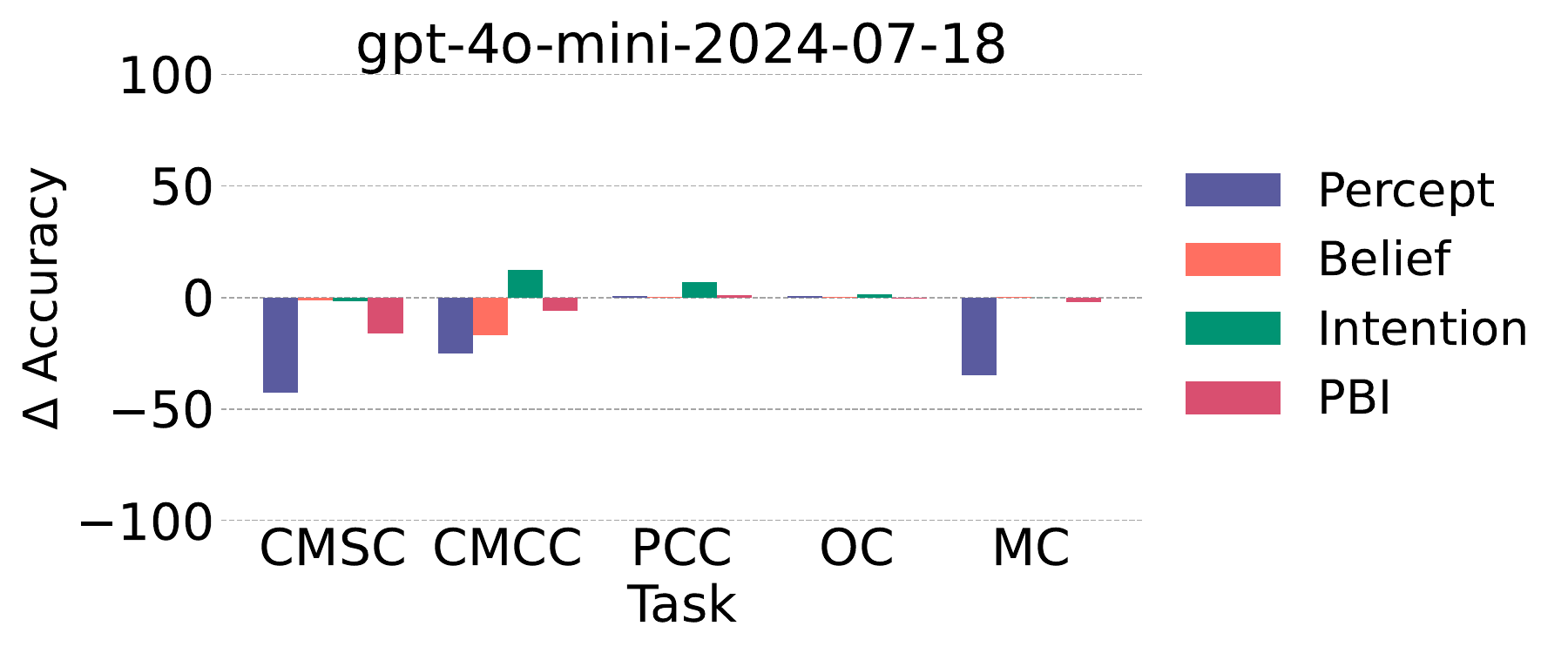}
    \end{subfigure}
    \hfill
    \begin{subfigure}[b]{0.48\textwidth}
        \centering
        \includegraphics[width=\linewidth]{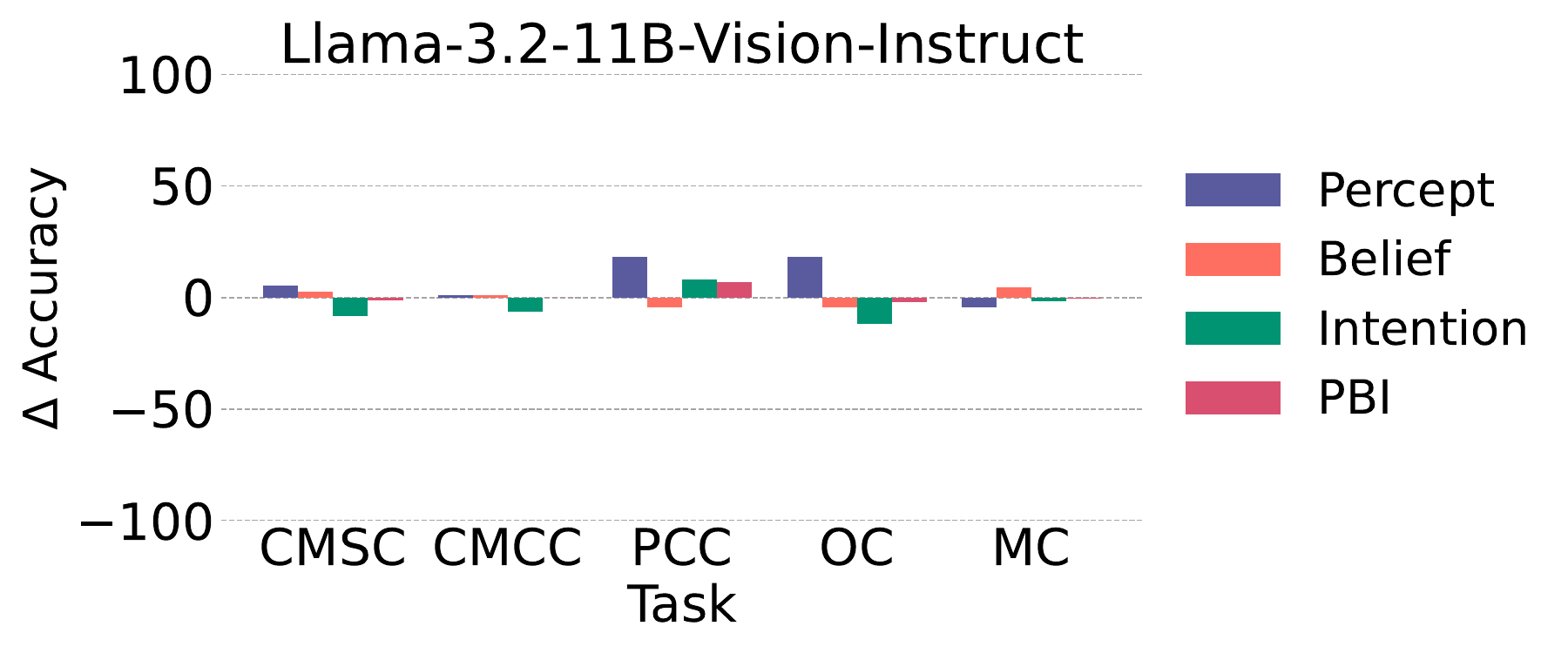}
    \end{subfigure}
    \hfill
    \begin{subfigure}[b]{0.48\textwidth}
        \centering
        \includegraphics[width=\linewidth]{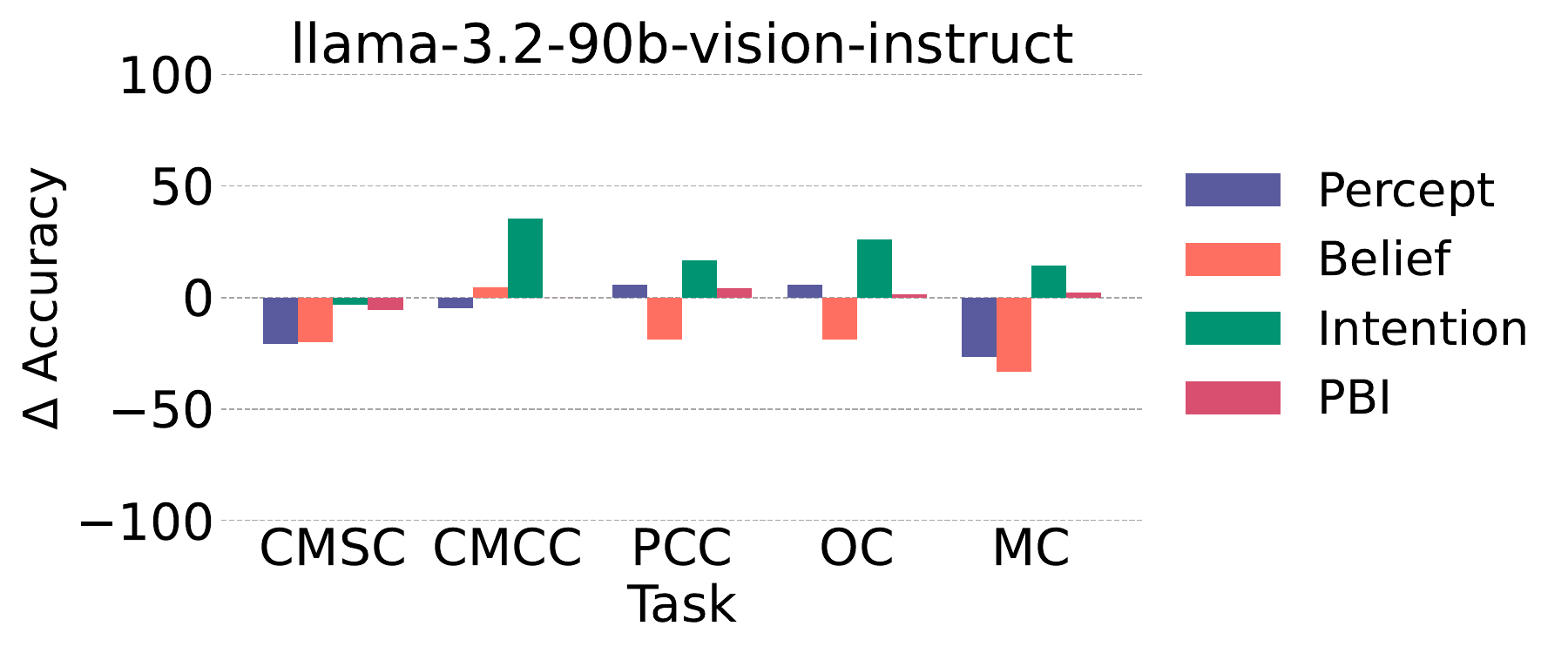}
    \end{subfigure}
    \hfill
    \begin{subfigure}[b]{0.48\textwidth}
        \centering
        \includegraphics[width=\linewidth]{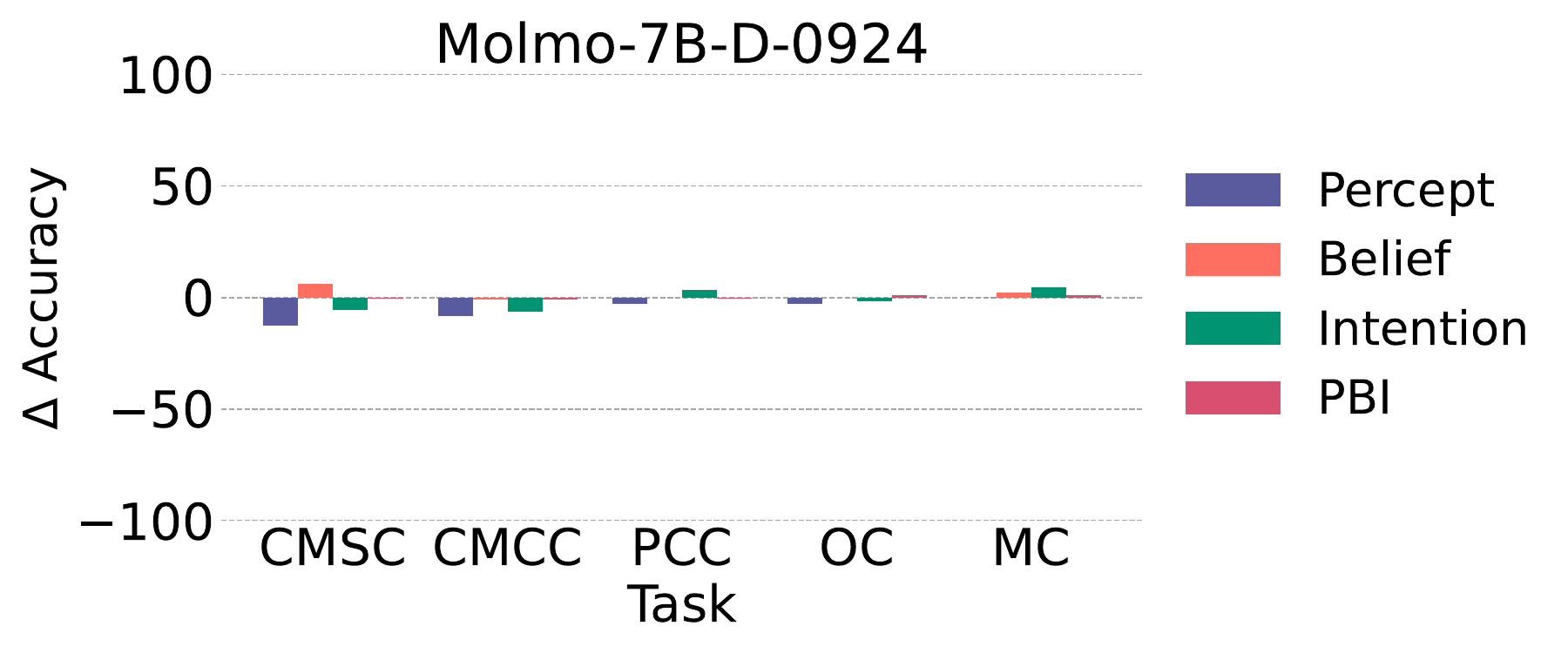}
    \end{subfigure}
    \hfill
    \begin{subfigure}[b]{0.48\textwidth}
        \centering
        \includegraphics[width=\linewidth]{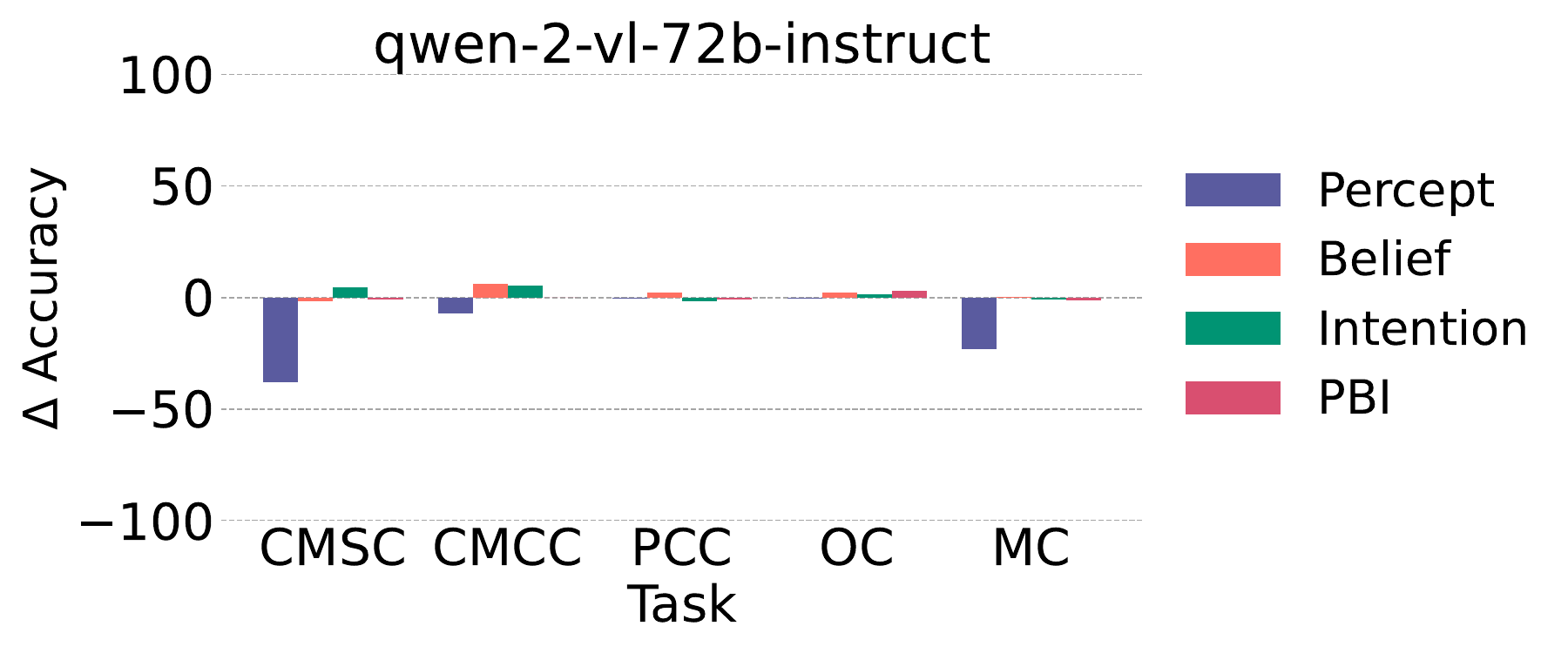}
    \end{subfigure}
    \caption{Difference in accuracy of VLMs when evaluated on the multimodal version of the \dataset{} versus the text-only version.}
    \label{fig:vl-diff-all}
\end{figure*}

\subsubsection{Gridworld Transformations}
\label{app:transform}
\dataset{} builds grid world environments starting from minial templates where agents are placed in pre-defined cells. 
Our generation pipeline then uniformly increases the distance between agents by a random value $\delta \in \{0, 1, 2, 3\}$ and applies one random transformation.
We present the formal definitions of transformations applied to our grid world templates, represented as a two-dimensional matrix $T \in \mathbb{R}^{m \times n}$, where $m$ is the number of rows and $n$ is the number of columns (in our specific case, $n=m$). Transformations include rotations, mirroring, and transposition, which are defined as follows.

\paragraph{90° Rotation}  
The 90° clockwise rotation of $T$ produces a new matrix $T'$ of size $n \times m$ such that:
\begin{equation}
T'_{i,j} = T_{m-j+1, i}, \quad \forall i \in [1, n], \, j \in [1, m].
\end{equation}
Operationally, this is equivalent to reversing the row order of $T$ and transposing:
\begin{equation}
T' = \text{Transpose}(\text{ReverseRows}(T)).
\end{equation}

\paragraph{180° Rotation}  
The 180° rotation produces a matrix $T'$ such that:
\begin{equation}
T'_{i,j} = T_{m-i+1, n-j+1}, \quad \forall i \in [1, m], \, j \in [1, n].
\end{equation}
This operation reverses both the rows and columns:
\begin{equation}
T' = \text{ReverseRows}(\text{ReverseColumns}(T)).
\end{equation}

\paragraph{270° Rotation}  
The 270° clockwise rotation produces a matrix $T'$ of size $n \times m$ such that:
\begin{equation}
T'_{i,j} = T_{j, n-i+1}, \quad \forall i \in [1, n], \, j \in [1, m].
\end{equation}
Operationally, this is equivalent to transposing $T$ and then reversing the rows:
\begin{equation}
T' = \text{ReverseRows}(\text{Transpose}(T)).
\end{equation}

\paragraph{Horizontal Mirroring}  
Horizontal mirroring reflects $T$ across its horizontal axis, producing a matrix $T'$ such that:
\begin{equation}
T'_{i,j} = T_{m-i+1, j}, \quad \forall i \in [1, m], \, j \in [1, n].
\end{equation}
This operation reverses the row order:
\begin{equation}
T' = \text{ReverseRows}(T).
\end{equation}

\paragraph{Vertical Mirroring}  
Vertical mirroring reflects $T$ across its vertical axis, producing a matrix $T'$ such that:
\begin{equation}
T'_{i,j} = T_{i, n-j+1}, \quad \forall i \in [1, m], \, j \in [1, n].
\end{equation}
This operation reverses the column order:
\begin{equation}
T' = \text{ReverseColumns}(T).
\end{equation}

\paragraph{Transposition}  
Transposition exchanges the rows and columns of $T$, producing a matrix $T'$ of size $n \times m$ such that:
\begin{equation}
T'_{i,j} = T_{j,i}, \quad \forall i \in [1, n], \, j \in [1, m].
\end{equation}
Operationally:
\begin{equation}
T' = \text{Transpose}(T).
\end{equation}

\begin{figure*}[t]
    \centering
    \includegraphics[width=\linewidth]{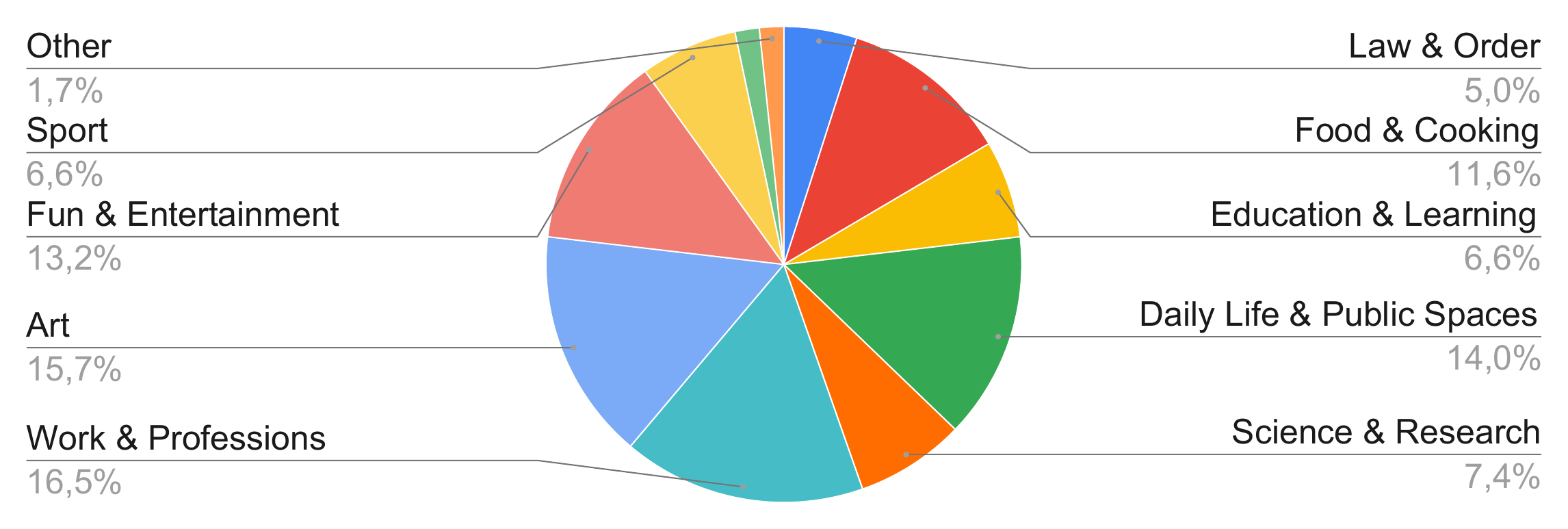}
    \caption{Distribution of the social contexts used in \dataset{}.}
    \label{fig:social-context-distr}
\end{figure*}

\subsection{Social Context Generation}
\label{app:stories}
To generate the social contexts used in our prompts, we employed a few-shot learning approach. 
We started by manually crafting five example social contexts, such as:
\texttt{a design studio where four graphic designers are working on a project}.
Social contexts are paired with four information IDs to use as agents' knowledge, for example \texttt{Feedback Loop}, \texttt{Color Scheme}, \texttt{Font Choice}, and \texttt{Design Concept}.
These examples were provided as input prompts to GPT-4o~\citep{gpt4o}, which we used iteratively to generate 115 additional social contexts. 
To ensure quality, we manually checked all generated contexts for consistency, meaningfulness, and to eliminate duplicates. 
Three people (native speakers or high proficiency) manually checked each social context. 
Additionally, we ran two rounds of a pilot study before finalising the dataset.
To compute the distribution of the generated social contexts, we initially asked two annotators to label them. 
We then measured the inter-annotator agreement using the Cohen's kappa, which was on the border between \textit{moderate} and \textit{substantial} agreement ($\kappa = 0.605$). 
Finally, the annotators proceeded to resolve disagreements. 
We report the final distribution of the social context topics in Figure~\ref{fig:social-context-distr}.

\begin{figure}[t]
    \centering
    \includegraphics[width=\linewidth]{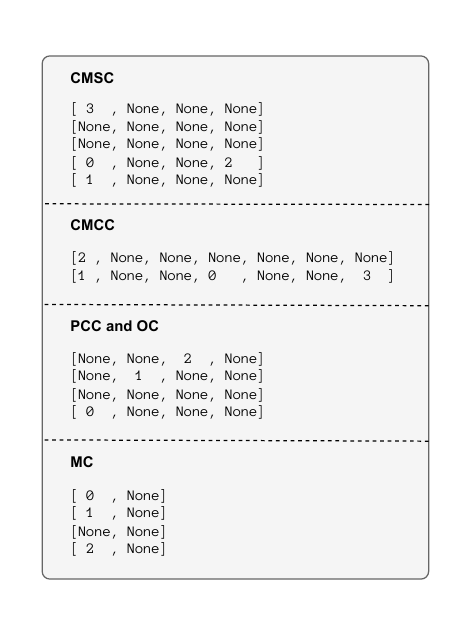}
    \caption{Minimal templates for each task. Agent IDs are denoted as integers (\texttt{1}, \texttt{2}, \texttt{3}, \texttt{4}) while \texttt{None} denotes an empty cell.}
    \label{fig:minimal-templates}
\end{figure}

\begin{algorithm}[t]
\caption{Generation pipeline}
\begin{algorithmic}[1]
\Require Set of agents $A \ni A_j$, $|A| = N_A$. Set of agents' initial knowledge $I_A \ni I_{A_j}$,  $j \in [0, N_A-1]$. Minimal grid template $T$ of size $s \times s$. Set of grid transformations $\mathcal{T}$. Set of social contexts $\mathcal{C}$. Prompt template $P$
\Require $\mathrm{isMultimodal} \in \{0,1\}$
\State $A' = \mathrm{RandomShuffle}(A)$
\State Sample $\delta \sim \mathrm{Uniform}(0, 3)$
\State $T' = \mathrm{IncreaseSize}(T, s')$, where $s' = s + \delta$
\State $T'' = f(T')$, where $f \sim \mathrm{\mathcal{T}}$
\If{$\mathrm{isMultimodal}$}
    \State $\mathrm{GridRepr} = \mathrm{RenderImage}(T'')$
\Else
    \State $\mathrm{GridRepr} = \mathrm{RenderText}(T'')$
\EndIf
\State Sample $c \sim \mathcal{C}$
\State $I'_A = \mathrm{RandomShuffle}(I_A)$
\State $\mathrm{Prompt} = \mathrm{MakePrompt}(P, A', c, I'_A, \mathrm{GridRepr})$
\State $\mathrm{Prompt'} = \mathrm{ShuffleAnswers}(\mathrm{Prompt})$
\State \Return $\mathrm{Prompt'}$, $\mathrm{GridRepr}$
\end{algorithmic}
\label{alg:generation}
\end{algorithm}

\begin{figure}[t]
    \centering
    \includegraphics[width=\linewidth]{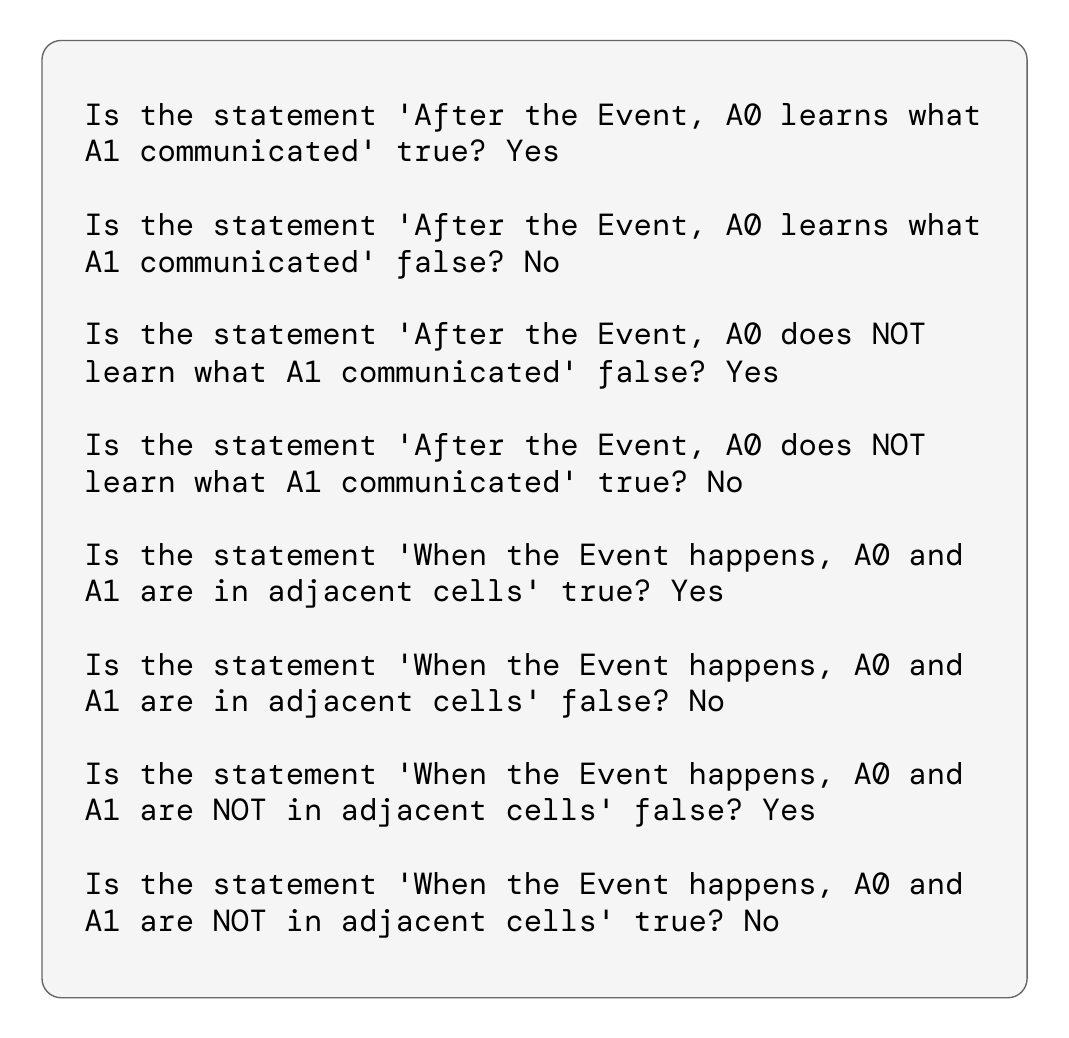}
    \caption{Example of question variants for the CMSC task.}
    \label{fig:question-variants}
\end{figure}

\begin{table}[t]
    \centering
    \resizebox{\linewidth}{!}{
    \begin{tabular}{lccc}
        \toprule
        \multirow{2}{*}{\textbf{Task}} & \multicolumn{3}{c}{\textbf{Question Type}} \\
        \cline{2-4}
         & \textbf{Percept} & \textbf{Belief} & \textbf{Intent} \\
        \midrule
        \textbf{CMSC} & $2132 \pm 71$ & $2188 \pm 81$ & $2126 \pm 73$ \\
        \textbf{CMCC} & $2266 \pm 86$ & $2567 \pm 102$ & $2241 \pm 87$ \\
        \textbf{PC} & $2134 \pm 71$ & $2240 \pm 86$ & $2188 \pm 80$ \\
        \textbf{OC} & $2134 \pm 71$ & $2240 \pm 85$ & $2134 \pm 79$ \\
        \textbf{MC} & $2114 \pm 68$ & $2175 \pm 77$ & $2172 \pm 77$ \\
        \bottomrule
    \end{tabular}
    }
    \caption{Average length and standard deviation of prompts across different tasks and question types.}
    \label{tab:lengths}
\end{table}

\subsection{Dataset Generation}
\label{app:prompts}
\dataset{} is entirely generated by code. 
The starting minimal template for each task is shown in Figure \ref{fig:minimal-templates}.
Each minimal template is paired with agents' initial knowledge and the correct answer, which is determined as discussed in \S\ref{sec:design} and \ref{app:utilityf}.
We show \dataset{}'s generation pipeline in detail in Algorithm~\ref{alg:generation},

The structure of prompts in \dataset{} is illustrated in Figure~\ref{fig:prompt-structure}, where we highlight the differences between the prompt used for VLMs (where the text representation of the grid is substituted by the image) and LLMs. 
The prompt first introduces the social context and information about agents.
It then presents the grid with the agents in their initial position as an image (for VLMs) or text (for LMs), as shown in Figure~\ref{fig:grid}.
Following the grid, the prompt lists the information initially known by the agents, the attitude of the target agent, and the event(s) that trigger a change in the environment. 
The prompt ends with the question and multiple-choice answers for the model to select from.

\dataset{} allows to define different formats for each question, as shown in Figure \ref{fig:question-variants}. 
During the dataset generation, the format is chosen randomly to avoid bias and make questions more diverse.
The correct answer to a question depends on multiple factors: the specific task being evaluated, 
the social context, the spatial setup of the environment, the attitude of the target agent, and the format of the question. 
The format of the question alone is not sufficient to answer the question. 
Previous work has often used just one single question format \citep{le2019revisiting, he2023hi, gandhi2024understanding, chan2024negotiationtom}. 
In the dataset version included with this submission, questions appear in two different formats.

We report the average number of characters per prompt for each task in Table \ref{tab:lengths}.
Examples of prompts for the three questions are reported in Example~\ref{box:percept} (percept), Example~\ref{box:belief} (belief), and Example~\ref{box:intent} (intent).

\subsection{Baseline Models}
\label{app:models}
We evaluate the following models: 
\begin{itemize}[leftmargin=15pt,topsep=5pt,itemsep=1pt,partopsep=1ex,parsep=1ex]
    \item \texttt{Llama-3.2-1B-Instruct} (unimodal)
    \item \texttt{Llama-3.2-3B-Instruct} (unimodal)
    \item \texttt{Llama-3.2-11B-Vision-Instruct} (uni- and multimodal settings) 
    \item \texttt{Llama-3.2-90B-Vision-Instruct} (uni- and multimodal settings) 
    \item \texttt{Mistral-7B-Instruct-v0.3} (unimodal)
    \item \texttt{Molmo-7B-D-0924} (uni- and multimodal)
    \item \texttt{Qwen2-VL-7B-Instruct} (uni- and multimodal)
    \item \texttt{Qwen2-VL-72B-Instruct} (uni- and multimodal)
    \item \texttt{gemma-2-9b-it} (unimodal)
    \item \texttt{claude-3.5-sonnet-20241022} (uni- and multimodal settings)
    \item \texttt{claude-3.5-haiku-20241022} (uni- and multimodal settings) 
    \item \texttt{gemini-pro-1.5} (uni- and multimodal settings) 
    \item \texttt{gemini-flash-1.5} (uni- and multimodal settings) 
    \item \texttt{gemini-2.5-flash} (uni- and multimodal settings) 
    \item \texttt{gpt-4o-2024-11-20} (uni- and multimodal settings) 
    \item \texttt{gpt-4o-mini-2024-07-18} (uni- and multimodal settings) 
    \item \texttt{o4-mini-2025-04-16} (multimodal settings) 
\end{itemize}
All models are used with a temperature of 0, to make them as deterministic as possible.

\subsection{Human Study}
\label{app:human-study}
We recruited 20 human participants (8 female, 12 male, aged between 21 and 40 years old) and asked them to answer 45 questions, equally split between tasks.
The study was approved by the institutional ethics committee.
Some participants were university students who received course credits as compensation, in accordance with university regulations.
The remaining participants voluntarily joined the study, without receiving any form of compensation.
Human participants were shown the same prompt used for evaluating models.
At the beginning of the study, participants were informed about their task, the duration of the experiment, and that their responses would be kept anonymous and used solely for research purposes.
They then went through a guided example that explains the rules of social interactions, so that they could optionally skip the first part of the prompt, which contains the same rules for each sample.
Between tasks, participants had to answer additional attention-check questions to ensure their attentiveness.

\subsection{Task Accuracy}
\label{app:fullscores}

Table~\ref{tab:results} contains the accuracy achieved by the baseline models on each question type -- Percept (P), Belief (B), Intention (I) -- and on PBI (P $\land$ B $\land$ I) for the five tasks in \dataset{}. 

Overall, models perform well—and in some cases even perfectly—when answering individual question types (P, B, or I). 
However, their performance drops when evaluated on the full PBI score, suggesting a lack of consistency across the different types of inference required to understand percepts, beliefs, and intentions.

In the text-only setting, Claude 3.5 Sonnet is the top-performing model across all tasks, except for CMCC, where Gemini Pro 1.5 performs better. 
When looking at the average PBI scores across models, the most challenging task is  CMCC ($4.6$), followed by CMSC ($12.8$), MC ($12.8$), OC ($13.9$), and finally PCC ($20.3$). 
Notably, there is a significant performance gap between CMCC and the other tasks. 
As shown in Table~\ref{tab:results}, this is largely due to low performance on  Belief questions, which are especially challenging as they require second-order ToM reasoning. 

In the multimodal setting, the best-performing models vary across tasks: Gemini Pro 1.5 leads on CMSC, Claude 3.5 Sonnet excels on CMCC and MC, and Qwen2-VL Instruct 7B performs best on PCC and OC. 
The trend in difficulty remains similar to the text-only setting, with CMCC again being the most challenging task by far ($0.93$).
Table~\ref{tab:results} shows again that this is due to a poor performance in Belief questions. 
The main difference in the multimodal setting is that MC becomes more difficult than CMSC, while PCC and OC continue to yield the highest average PBI scores across models.

\subsection{P, PB, and PBI Inference}
\label{app:pbi-chain}
Figure~\ref{fig:chain-all}a shows the change in performance across P, PB, and PBI accuracy for all the VLMs we evaluated on \dataset{}.
Similarly, Figure~\ref{fig:chain-all}b shows the performance of LMs on the text-only version of \dataset{}. 
Overall, model performance generally declines from percepts to beliefs to intentions. 
However, LMs evaluated on text demonstrate greater robustness in PB inference across most tasks (CMSC, PCC, OC, MC). 
Among them, the top-performing models, Claude 3.5 Sonnet and GPT-4o, achieve PB accuracy comparable to human levels in CMSC and MC.

\subsection{Multimodal vs Unimodal Performance}
\label{app:l-vs-vl}
In \S\ref{sec:results-analyses} we show that GPT-4o does not benefit from the addition of images (Figure~\ref{fig:vl-diff}, top), while Qwen2 shows a clear benefit from the image input, particularly on percept questions (Figure~\ref{fig:vl-diff}, bottom). 
Figure~\ref{fig:vl-diff-all} shows additional comparisons of VLMs' performance when evaluated on the multimodal version \dataset{} versus the text-only version.
In general, models do not particularly benefit from images, especially Claude~3.5~Sonnet, Gemini-1.5-Pro, and GPT-4o.
For other models like Llama~3.2 and Molmo, performance does not change much between modalities.  
Figure~\ref{fig:vl-diff-all} presents additional comparisons of VLM performance on the multimodal version of \dataset{} versus the text-only version. 
Overall, state-of-the-art models -- Claude, Gemini and GPT -- do not benefit from images and show stronger performance on text. 
Other models, like Llama 3.2 and Molmo, show some (but not consistent) improvements when evaluated in the multimodal setting.
Gemini~2.5~Flash also benefits from multimodal inputs, especially for Belief questions in PCC, OC, and to a lesser extent in MC.

\paragraph{The Strange Case of GPT-4o's Coordinate System} 
As discussed in \S\ref{sec:results-analyses}, \S\ref{app:l-vs-vl}, and shown in detail in Table~\ref{tab:results}, GPT-4o's performance on multimodal percept questions is poor across all tasks, which in turn results in a low PBI accuracy. 
To gain more insight, we examined GPT-4o's generated output and found that the model often attempts to answer percept questions by defining a coordinate system for the grid world. 
In such cases, two issues arise: first, the coordinate system defined by GPT-4o changes across different samples. 
Second, even within a single sample and using its own coordinate system, the model often fails to correctly position agents on the grid. 
In contrast, GPT-4o demonstrates higher accuracy when it reasons in terms of relative positions, such as ``$A_j$ is to the right of $A_k$''.

\subsection{Compute resources}
We ran open-source models of size below 15B on a server running Ubuntu 22.04, equipped with eight NVIDIA Tesla V100-SXM2 GPUs with 32GB of memory and Intel Xeon Platinum 8260 CPUs. 
Proprietary models are used through API.

\subsection{Code}
Our code is public under the MIT license at \url{https://git.hcics.simtech.uni-stuttgart.de/public-projects/tom-ssi}.

\onecolumn
\begin{example}{box:percept}{PerceptColor}{Percept prompt}
\begin{lstlisting}[basicstyle=\ttfamily\scriptsize, breaklines=true]
Task:
You are a helpful AI assistant tasked to answer a question about a designer in a design studio.

Setup:
You are observing a gridworld that represents a design studio where 4 graphic designers are working on a project.

Rules:
1. The design studio is represented as a grid of size 6x6 with 4 designers: A0, A1, A2, and A3. Each designer occupies one cell.
2. Each designer starts out knowing a subset of 4 pieces of information: ['Feedback Loop', 'Color Scheme', 'Font Choice', 'Design Concept'].
3. By default, designers have two goals: (1) learning information they are missing from other designers, and (2) communicating information that other designers are missing. Setting a designer's Attitude can influence their goals.
4. Each designer can see the positions, movements, and initial information of other designers. However, they cannot see any new information other designers learn later, so they must infer it.
5. Event(s) change the state of the design studio and include: movement of designers (up, down, left, right) or communication of information. Aside from what is specified in the Event(s), no other movement or communication happens.
6. If a designer shares a piece of information during the Event(s), only the designers in the adjacent cells (up, down, left, right, and diagonally neighboring cells) will learn it. 

Initial position of the designers in the design studio:
+----+----+----+----+----+----+
|    |    |    |    | A1 |    |
+----+----+----+----+----+----+
| A3 |    |    |    | A2 |    |
+----+----+----+----+----+----+
|    |    |    |    |    |    |
+----+----+----+----+----+----+
|    |    |    |    |    |    |
+----+----+----+----+----+----+
|    |    |    |    |    |    |
+----+----+----+----+----+----+
|    |    |    |    | A0 |    |
+----+----+----+----+----+----+

Information initially known by the designers:
- A0: ['Font Choice', 'Design Concept']
- A1: ['Feedback Loop', 'Design Concept']
- A2: ['Color Scheme', 'Design Concept']
- A3: ['Feedback Loop', 'Color Scheme', 'Font Choice']

A2's Attitude:
Collaborative: A2 wants to learn new information from other designers and share information that other designers don't know. 

Event:
Designer A1 communicates 'Feedback Loop'.

Question:
Is the statement 'After the Event, A2 learns what A1 communicated' true?

Based on the information above, choose one of the following answers:
A) No
B) Yes

Provide your final answer within the tags <Answer>[answer]</Answer> (e.g. <Answer>A</Answer>).
\end{lstlisting}
\end{example}
\twocolumn
\onecolumn
\begin{example}{box:belief}{BeliefColor}{Belief prompt}
\begin{lstlisting}[basicstyle=\ttfamily\scriptsize, breaklines=true]
Task:
You are a helpful AI assistant tasked to answer a question about a designer in a design studio.

Setup:
You are observing a gridworld that represents a design studio where 4 graphic designers are working on a project.

Rules:
1. The design studio is represented as a grid of size 6x6 with 4 designers: A0, A1, A2, and A3. Each designer occupies one cell.
2. Each designer starts out knowing a subset of 4 pieces of information: ['Feedback Loop', 'Color Scheme', 'Font Choice', 'Design Concept'].
3. By default, designers have two goals: (1) learning information they are missing from other designers, and (2) communicating information that other designers are missing. Setting a designer's Attitude can influence their goals.
4. Each designer can see the positions, movements, and initial information of other designers. However, they cannot see any new information other designers learn later, so they must infer it.
5. Event(s) change the state of the design studio and include: movement of designers (up, down, left, right) or communication of information. Aside from what is specified in the Event(s), no other movement or communication happens.
6. If a designer shares a piece of information during the Event(s), only the designers in the adjacent cells (up, down, left, right, and diagonally neighboring cells) will learn it. 

Initial position of the designers in the design studio:
+----+----+----+----+----+----+
|    |    |    |    | A1 |    |
+----+----+----+----+----+----+
| A3 |    |    |    | A2 |    |
+----+----+----+----+----+----+
|    |    |    |    |    |    |
+----+----+----+----+----+----+
|    |    |    |    |    |    |
+----+----+----+----+----+----+
|    |    |    |    |    |    |
+----+----+----+----+----+----+
|    |    |    |    | A0 |    |
+----+----+----+----+----+----+

Information initially known by the designers:
- A0: ['Font Choice', 'Design Concept']
- A1: ['Feedback Loop', 'Design Concept']
- A2: ['Color Scheme', 'Design Concept']
- A3: ['Feedback Loop', 'Color Scheme', 'Font Choice']

A2's Attitude:
Collaborative: A2 wants to learn new information from other designers and share information that other designers don't know. 

Event:
Designer A1 communicates 'Feedback Loop'.

Question:
After the Event, what information does A2 believe she is still missing?

Based on the information above, choose one of the following answers:
A) Design Concept
B) Feedback Loop
C) Font Choice
D) Color Scheme

Provide your final answer within the tags <Answer>[answer]</Answer> (e.g. <Answer>A</Answer>).
\end{lstlisting}
\end{example}
\twocolumn
\onecolumn
\begin{example}{box:intent}{IntentColor}{Intention prompt}
\begin{lstlisting}[basicstyle=\ttfamily\scriptsize, breaklines=true]
Task:
You are a helpful AI assistant tasked to answer a question about a designer in a design studio.

Setup:
You are observing a gridworld that represents a design studio where 4 graphic designers are working on a project.

Rules:
1. The design studio is represented as a grid of size 6x6 with 4 designers: A0, A1, A2, and A3. Each designer occupies one cell.
2. Each designer starts out knowing a subset of 4 pieces of information: ['Feedback Loop', 'Color Scheme', 'Font Choice', 'Design Concept'].
3. By default, designers have two goals: (1) learning information they are missing from other designers, and (2) communicating information that other designers are missing. Setting a designer's Attitude can influence their goals.
4. Each designer can see the positions, movements, and initial information of other designers. However, they cannot see any new information other designers learn later, so they must infer it.
5. Event(s) change the state of the design studio and include: movement of designers (up, down, left, right) or communication of information. Aside from what is specified in the Event(s), no other movement or communication happens.
6. If a designer shares a piece of information during the Event(s), only the designers in the adjacent cells (up, down, left, right, and diagonally neighboring cells) will learn it. 

Initial position of the designers in the design studio:
+----+----+----+----+----+----+
|    |    |    |    | A1 |    |
+----+----+----+----+----+----+
| A3 |    |    |    | A2 |    |
+----+----+----+----+----+----+
|    |    |    |    |    |    |
+----+----+----+----+----+----+
|    |    |    |    |    |    |
+----+----+----+----+----+----+
|    |    |    |    |    |    |
+----+----+----+----+----+----+
|    |    |    |    | A0 |    |
+----+----+----+----+----+----+

Information initially known by the designers:
- A0: ['Font Choice', 'Design Concept']
- A1: ['Feedback Loop', 'Design Concept']
- A2: ['Color Scheme', 'Design Concept']
- A3: ['Feedback Loop', 'Color Scheme', 'Font Choice']

A2's Attitude:
Collaborative: A2 wants to learn new information from other designers and share information that other designers don't know. 

Event:
Designer A1 communicates 'Feedback Loop'.

Question:
After the Event, which designer is A2 most likely to approach?

Based on the information above, choose one of the following answers:
A) A3
B) A0
C) A1

Provide your final answer within the tags <Answer>[answer]</Answer> (e.g. <Answer>A</Answer>).
\end{lstlisting}
\end{example}
\twocolumn
\onecolumn
\begin{example}{box:llama-cmsc-pb-error}{BlueViolet}{PB Error Case in CMSC -- Llama-3.2-11B-Vision}
\begin{lstlisting}[basicstyle=\ttfamily\scriptsize, breaklines=true, escapechar={@}]
Task:
You are a helpful AI assistant tasked to answer a question about a reporter in a newsroom.

Setup:
You are observing a gridworld that represents a newsroom where 4 reporters are gathering information for breaking news.

Rules:
1. The newsroom is represented as a grid of size 11x11 with 4 reporters: A0, A1, A2, and A3. Each reporter occupies one cell.
2. Each reporter starts out knowing a subset of 4 pieces of information: ['Breaking News', 'Press Release', 'Eyewitness Account', 'Photograph'].
3. By default, reporters have two goals: (1) learning information they are missing from other reporters, and (2) communicating information that other reporters are missing. Setting a reporter's Attitude can influence their goals.
4. Each reporter can see the positions, movements, and initial information of other reporters. However, they cannot see any new information other reporters learn later, so they must infer it.
5. Event(s) change the state of the newsroom and include: movement of reporters (up, down, left, right) or communication of information. Aside from what is specified in the Event(s), no other movement or communication happens.
6. If a reporter shares a piece of information during the Event(s), only the reporters in the adjacent cells (up, down, left, right, and diagonally neighboring cells) will learn it.

The initial position of the reporters in the newsroom is shown in the provided image.

Information initially known by the reporters:
- A0: ['Breaking News', 'Press Release', 'Eyewitness Account']
- A1: ['Breaking News', 'Press Release', 'Eyewitness Account']
- A2: ['Breaking News', 'Press Release', 'Eyewitness Account', 'Photograph']
- A3: ['Breaking News', 'Eyewitness Account', 'Photograph']

A2's Attitude:
Collaborative: A2 wants to learn new information from other reporters and share information that other reporters don't know.

Events:
Reporter A3 communicates 'Photograph'.
Reporter A1 communicates 'Press Release'.

Question:
After the Event, which information does A2 believe A3 and A1 are likely still missing, respectively?

Based on the information above and on the image, choose one of the following answers:
A) A2 believes that A3 is missing Breaking News and A1 is missing Press Release
B) A2 believes that A3 and A1 are not missing any information
C) A2 believes that A3 is missing Eyewitness Account and A1 is missing Photograph
D) A2 believes that A3 is missing Press Release and A1 is missing Photograph

Provide your final answer within the tags <Answer>[answer]</Answer> (e.g. <Answer>A</Answer>).

@\textcolor{Red}{To determine which information A2 believes A3 and A1 are likely still missing, we need to analyze the events that have occurred and the initial information known by each reporter.}@

@\textcolor{Red}{Initially, A2 knows all four pieces of information: ['Breaking News', 'Press Release', 'Eyewitness Account', 'Photograph']. A3 knows ['Breaking News', 'Eyewitness Account', 'Photograph'], and A1 knows ['Breaking News', 'Press Release', 'Eyewitness Account'].}@

@\textcolor{Red}{During the events, Reporter A3 communicates 'Photograph' and Reporter A1 communicates 'Press Release'.}@

@\textcolor{Red}{After these events, A2 will update its beliefs about what A3 and A1 are likely still missing.}@

@\textcolor{Red}{A2 believes that A3 is missing Press Release because A3 did not communicate it, and A2 knows that A3 knows 'Photograph' but not 'Press Release'. A2 believes that A1 is missing Photograph because A1 did not communicate it, and A2 knows that A1 knows 'Press Release' but not 'Photograph'.}@

@\textcolor{Red}{Therefore, the correct answer is:}@

@\textcolor{Red}{<Answer>D</Answer>}@
\end{lstlisting}

\tcblower

Correct answer: B

\end{example}
\twocolumn
\onecolumn
\begin{example}{box:llama-cmsc-pbi-error}{BlueViolet}{PBI Error Case in CMSC -- Llama-3.2-11B-Vision}
\begin{lstlisting}[basicstyle=\ttfamily\scriptsize, breaklines=true, escapechar={@}]
Task:
You are a helpful AI assistant tasked to answer a question about a baker in a bake sale.

Setup:
You are observing a gridworld that represents a charity bake sale where 4 bakers are collaborating to raise funds.

Rules:
1. The bake sale is represented as a grid of size 7x7 with 4 bakers: A0, A1, A2, and A3. Each baker occupies one cell.
2. Each baker starts out knowing a subset of 4 pieces of information: ['Special Ingredient', 'Customer Feedback', 'Baked Good Description', 'Price Tag'].
3. By default, bakers have two goals: (1) learning information they are missing from other bakers, and (2) communicating information that other bakers are missing. Setting a baker's Attitude can influence their goals.
4. Each baker can see the positions, movements, and initial information of other bakers. However, they cannot see any new information other bakers learn later, so they must infer it.
5. Event(s) change the state of the bake sale and include: movement of bakers (up, down, left, right) or communication of information. Aside from what is specified in the Event(s), no other movement or communication happens.
6. If a baker shares a piece of information during the Event(s), only the bakers in the adjacent cells (up, down, left, right, and diagonally neighboring cells) will learn it.

The initial position of the bakers in the bake sale is shown in the provided image.

Information initially known by the bakers:
- A0: ['Special Ingredient', 'Customer Feedback', 'Baked Good Description']
- A1: ['Special Ingredient', 'Baked Good Description', 'Price Tag']
- A2: ['Special Ingredient', 'Customer Feedback', 'Baked Good Description', 'Price Tag']
- A3: ['Special Ingredient', 'Customer Feedback', 'Baked Good Description']

A2's Attitude:
Collaborative: A2 wants to learn new information from other bakers and share information that other bakers don't know.

Events:
Baker A1 communicates 'Price Tag'.
Baker A3 communicates 'Customer Feedback'.

Question:
After the Events, which baker is A2 most likely to approach?

Based on the information above and on the image, choose one of the following answers:
A) A0
B) A1
C) A3

Provide your final answer within the tags <Answer>[answer]</Answer> (e.g. <Answer>A</Answer>).

@\textcolor{Red}{To determine which baker A2 is most likely to approach after the events, we need to analyze the information shared by A1 and A3.}@

@\textcolor{Red}{Initially, A2 knows 'Special Ingredient', 'Customer Feedback', and 'Baked Good Description'. After the events, A1 communicates 'Price Tag', and A3 communicates 'Customer Feedback'. Since A2 already knows 'Customer Feedback', the new information shared by A3 is not relevant to A2's current knowledge.}@

@\textcolor{Red}{However, A1's communication of 'Price Tag' is new information for A2. As A2 has a collaborative attitude, she wants to learn new information from other bakers. Therefore, A2 is most likely to approach A1 to learn the 'Price Tag' information.}@

@\textcolor{Red}{**Answer:** B) A1}@
\end{lstlisting}

\tcblower

Correct answer: A

\end{example}
\twocolumn
\onecolumn
\begin{example}{box:llama-mc-pbi-error}{BlueViolet}{PBI Error Case in MC -- Llama-3.2-11B-Vision}
\begin{lstlisting}[basicstyle=\ttfamily\scriptsize, breaklines=true, escapechar={@}]
Task:
You are a helpful AI assistant tasked to answer a question about an artist in a art gallery.

Setup:
You are observing a gridworld that represents a local art gallery opening where 3 artists are showcasing their work.

Rules:
1. The art gallery is represented as a grid of size 7x7 with 3 artists: A0, A1, and A2. Each artist occupies one cell.
2. Each artist starts out knowing a subset of 4 pieces of information: ['Material Source', 'Art Piece Title', 'Technique Explanation', 'Inspiration Story'].
3. By default, artists have two goals: (1) learning information they are missing from other artists, and (2) communicating information that other artists are missing. Setting an artist's Attitude can influence their goals.
4. Each artist can see the positions, movements, and initial information of other artists. However, they cannot see any new information other artists learn later, so they must infer it.
5. Event(s) change the state of the art gallery and include: movement of artists (up, down, left, right) or communication of information. Aside from what is specified in the Event(s), no other movement or communication happens.
6. If an artist shares a piece of information during the Event(s), only the artists in the adjacent cells (up, down, left, right, and diagonally neighboring cells) will learn it.

The initial position of the artists in the art gallery is shown in the provided image.

Information initially known by the artists:
- A0: ['Art Piece Title', 'Technique Explanation', 'Inspiration Story']
- A1: ['Technique Explanation', 'Inspiration Story']
- A2: ['Material Source', 'Art Piece Title', 'Technique Explanation', 'Inspiration Story']

A2's Attitude:
Collaborative towards A1: A2 wants A1 to learn new information.
Obstructive towards A0: A2 doesn't want A0 to learn new information.

Event:
Artist A0 moves one cell down.

Question:
After the Event, which information is A2 most likely to communicate?

Based on the information above and on the image, choose one of the following answers:
A) Technique Explanation
B) Material Source
C) Art Piece Title
D) Inspiration Story

Provide your final answer within the tags <Answer>[answer]</Answer> (e.g. <Answer>A</Answer>).

@\textcolor{Red}{To determine which information A2 is most likely to communicate after the event, we need to analyze A2's Attitude and the initial information known by each artist.}@

@\textcolor{Red}{A2's Attitude is collaborative towards A1 and obstructive towards A0. This means A2 wants A1 to learn new information but does not want A0 to learn new information.}@

@\textcolor{Red}{Initially, A2 knows all the information except 'Material Source'. Since A2 is obstructive towards A0, it will not share any information with A0.}@

@\textcolor{Red}{A0 has moved one cell down, which does not change the fact that A2 is obstructive towards A0.}@

@\textcolor{Red}{Therefore, A2 will not share any information with A0.}@

@\textcolor{Red}{A2 is collaborative towards A1, but A1 already knows 'Technique Explanation' and 'Inspiration Story'. A2 knows 'Material Source' and 'Art Piece Title', but A1 already knows 'Art Piece Title'.}@

@\textcolor{Red}{A2 will share the information that A1 does not know.}@

@\textcolor{Red}{The information that A2 knows and A1 does not know is 'Material Source'.}@

@\textcolor{Red}{So, after the event, A2 is most likely to communicate 'Material Source'.}@

@\textcolor{Red}{<Answer>B</Answer>}@
\end{lstlisting}

\tcblower

Correct answer: C

\end{example}
\twocolumn
\onecolumn
\begin{example}{box:llama-mc-pbi-correct}{BlueViolet}{PBI Correct Case in MC -- Llama-3.2-11B-Vision}
\begin{lstlisting}[basicstyle=\ttfamily\scriptsize, breaklines=true, escapechar={@}]
Task:
You are a helpful AI assistant tasked to answer a question about a speaker in a tech conference.

Setup:
You are observing a gridworld that represents a tech conference where 3 speakers are presenting their innovations.

Rules:
1. The tech conference is represented as a grid of size 5x5 with 3 speakers: A0, A1, and A2. Each speaker occupies one cell.
2. Each speaker starts out knowing a subset of 4 pieces of information: ['Future Trends', 'Q&A Session', 'Technical Challenges', 'Innovation Overview'].
3. By default, speakers have two goals: (1) learning information they are missing from other speakers, and (2) communicating information that other speakers are missing. Setting a speaker's Attitude can influence their goals.
4. Each speaker can see the positions, movements, and initial information of other speakers. However, they cannot see any new information other speakers learn later, so they must infer it.
5. Event(s) change the state of the tech conference and include: movement of speakers (up, down, left, right) or communication of information. Aside from what is specified in the Event(s), no other movement or communication happens.
6. If a speaker shares a piece of information during the Event(s), only the speakers in the adjacent cells (up, down, left, right, and diagonally neighboring cells) will learn it.

The initial position of the speakers in the tech conference is shown in the provided image.

Information initially known by the speakers:
- A0: ['Q&A Session', 'Technical Challenges', 'Innovation Overview']
- A1: ['Technical Challenges', 'Innovation Overview']
- A2: ['Future Trends', 'Q&A Session', 'Technical Challenges', 'Innovation Overview']

A2's Attitude:
Collaborative towards A1: A2 wants A1 to learn new information.
Obstructive towards A0: A2 doesn't want A0 to learn new information.

Event:
Speaker A0 moves one cell left.

Question:
After the Event, which information is A2 most likely to communicate?

Based on the information above and on the image, choose one of the following answers:
A) Q&A Session
B) Innovation Overview
C) Future Trends
D) Technical Challenges

Provide your final answer within the tags <Answer>[answer]</Answer> (e.g. <Answer>A</Answer>).

@\textcolor{Green}{To determine which information A2 is most likely to communicate after the event, we need to consider A2's Attitude and the new positions of the speakers.}@

@\textcolor{Green}{A2's Attitude is Collaborative towards A1 and Obstructive towards A0. This means A2 wants to help A1 learn new information but does not want to help A0 learn new information.}@

@\textcolor{Green}{After the event, A0 moves one cell left, so the new positions are:}@

@\textcolor{Green}{* A0: (2, 1)}@
@\textcolor{Green}{* A1: (2, 2)}@
@\textcolor{Green}{* A2: (2, 3)}@

@\textcolor{Green}{Since A2 is obstructive towards A0, it is unlikely to communicate any information to A0. Therefore, we can rule out options that would benefit A0.}@

@\textcolor{Green}{A2 is collaborative towards A1, so it is likely to communicate information that A1 is missing. A1 initially knows 'Technical Challenges' and 'Innovation Overview', but it does not know 'Q\&A Session'. Since A2 is in a position where it can communicate with A1, it is most likely to communicate 'Q\&A Session', which is the only information A1 is missing.}@

@\textcolor{Green}{Therefore, the correct answer is:}@

@\textcolor{Green}{<Answer>A</Answer>}@
\end{lstlisting}

\tcblower

Correct answer: A

\end{example}
\twocolumn

\end{document}